\documentclass{IEEEtran}
\usepackage{cite,graphicx}
\usepackage{amsfonts,amssymb}
\usepackage{fancyhdr}
\usepackage{mdwmath}
\usepackage{mdwtab}
\usepackage{balance}
\usepackage{xcolor}
\usepackage{bm}
\usepackage{subfig}
\usepackage{amsmath}
\usepackage[ruled,linesnumbered]{algorithm2e}
\usepackage{algorithmic}
\usepackage{multirow}
\usepackage{flafter}
\usepackage{mathrsfs}
\usepackage{url}
\usepackage{multicol}
\usepackage{hyperref}
\usepackage{hhline}
\usepackage{makecell}
\usepackage[left= 0.625in, right= 0.625in,top=0.76in, bottom=0.97in]{geometry}

\begin{document}

\title{
Maximizing Uncertainty for Federated learning via Bayesian Optimisation-based Model Poisoning 
}

\author{
Marios Aristodemou,~\IEEEmembership{Graduate Student Member,~IEEE,}
Xiaolan Liu,~\IEEEmembership{Member,~IEEE,}
Yuan Wang, ~\IEEEmembership{Member,~IEEE,}
Konstantinos G. Kyriakopoulos,~\IEEEmembership{Senior Member,~IEEE,}
Sangarapillai Lambotharan,~\IEEEmembership{Senior Member,~IEEE,}
Qingsong Wei, ~\IEEEmembership{Senior Member,~IEEE,}
\thanks{Marios Aristodemou, and Konstantinos G. Kyriakopoulos are  with the Wolfson School of Mechanical, Electrical and Manufacturing Engineering, Loughborough University, Loughborough, U.K. }
\thanks{Xiaolan Liu is with Smart Internet Lab, University of Bristol, Bristol, U.K.}
\thanks{Sangarapillai Lambotharan is with the Institute for Digital Technologies, Loughborough University London, London, U.K.}
\thanks{Qingsong Wei, Yuan Wang is with the Department of Computing and Intelligence, Institute of High Performance Computing, under A*STAR, Singapore.}
\thanks{The emails for the corresponding authors are m.aristodemou@lboro.ac.uk and xiaolan.liu@bristol.ac.uk}
\thanks{This work was partially supported by the UK Engineering and Physical Sciences Research Council (EPSRC) under Grant EP/X012301/1, EP/X04047X/2 and EP/Y037243/1, by the Royal Society Research Grant, RG/R2/232525 ”Machine Learning enabled Efficient Communication Scheme for Metaverse over Wireless Networks”, and by the RIE2025 Industry Alignment Fund – Industry Collaboration Project (IAF-ICP) (Award No: I2301E0020), administered by A*STAR.}
\thanks{For the purpose of open access, the authors have applied a Creative Commons Attribution (CC-BY) license to any Author Accepted Manuscript version arising.}
}

\markboth{Transactions on Information on Forensics and Security, Accepted for publication}%
{Maximizing Uncertainty for Federated learning via Bayesian Optimisation-based Model Poisoning }

\maketitle

\begin{abstract}
As we transition from Narrow Artificial Intelligence towards Artificial Super Intelligence, users are increasingly concerned about their privacy and the trustworthiness of machine learning (ML) technology. A common denominator for the metrics of trustworthiness is the quantification of uncertainty inherent in DL algorithms, and specifically in the model parameters, input data, and model predictions.
One of the common approaches to address privacy-related issues in DL is to adopt distributed learning such as federated learning (FL), where private raw data is not shared among users. Despite the privacy-preserving mechanisms in FL, it still faces challenges in trustworthiness. Specifically, the malicious users, during training, can systematically create malicious model parameters to compromise the models’ predictive and generative capabilities, resulting in high uncertainty about their reliability.
To demonstrate malicious behaviour, we propose a novel model poisoning attack method named Delphi \footnote{Delphi is a temple located in the Ancient Greece, where a lot of leaders used to seek for advice. The oracles used to consult the leaders with two possible advices, which both were luckily to be happened resulting to high uncertainty on leaders' decision. Thus, we have named after the attack, because our methodology aims to provide a model which has high uncertainty like the advices from Delphi's oracle.} which aims to maximise the uncertainty of the global model output. We achieve this by taking advantage of the relationship between the uncertainty and the model parameters of the first hidden layer of the local model. Delphi employs two types of optimisation, Bayesian Optimisation and Least Squares Trust Region, to search for the optimal poisoned model parameters, named as Delphi-BO and Delphi-LSTR. We quantify the uncertainty using the KL Divergence to minimise the distance of the predictive probability distribution towards an uncertain distribution of model output.  Furthermore, we establish a mathematical proof for the attack effectiveness demonstrated in FL. Numerical results demonstrate that Delphi-BO induces a higher amount of uncertainty than Delphi-LSTR highlighting vulnerability of FL systems to model poisoning attacks.

\end{abstract}


\section{Introduction}
The advancement of AI needs to address user concerns regarding privacy and the trustworthiness of algorithms \cite{Zorpette_2023}. A comprehensive framework to evaluate attacks that compromise the reliability of these algorithms is thus essential. With the enforcement of the European Union’s AI Act \cite{ai_eu_2024}, AI algorithms are regulated based on the criticality of their applications and the risks posed by their predictions, making trustworthiness a key evaluation criterion.

Trustworthiness in AI models encompasses several aspects, such as explainability, interoperability, and accuracy \cite{Albahri2023}, all crucially tied to the quantification of uncertainty. This uncertainty, stemming from model parameters, input data, and prediction outcomes, significantly influences the confidence level in model predictions \cite{seedat2019calibrated}. By analyzing both epistemic uncertainty (model parameters) and aleatoric uncertainty (input data), we can gauge the uncertainty of the model and reveal the reliability of model predictions and the quality of training data \cite{smith2018understanding, Begoli201920}.

Beyond model reliability, privacy protection remains a paramount concern, particularly in sectors like healthcare where institutions handle sensitive patient data. Federated Learning (FL) offers a viable solution by enabling the development of generalizable models without direct data sharing among parties. Through FL, multiple data owners, under the coordination of a central server, collaboratively train models while keeping the raw data localized, thereby enhancing both security and privacy \cite{Lim2020, Tan2022, wang2024}. FL's utility spans various domains, including healthcare \cite{Chen2020}, resource allocation \cite{XiaolanHSFL2023}, and digital twins \cite{Lu2021}, showcasing its broad applicability.

\IEEEpubidadjcol

However, the deployment of deep learning also introduces the risk of adversarial attacks \cite{BIGGIO2018317, aristodemou2022, beechey2023}.L systems are particularly vulnerable to such threats, especially when malicious participants are involved in the learning process \cite{Aristodemou2023}. Such attacks aim to manipulate the local or global model parameters to cause erroneous predictions, low confidence in the model output \cite{fang2020local}, and data reconstruction \cite{shokri2017membership}. An example of how critical a poisoning attack in FL is the poisoned model parameters in an Internet of Things (IoT) network can compromise the trusted entities in FL and adversaries could gain control of the production chain \cite{rehman2021}. Adversaries may even gain critical information about the network by getting access to the devices. Another critical issue is that the adversaries gain the users' privacy-sensitive data (e.g., patients' data and social media account data), such as in Metaverse applications. Distributed learning is a driving force for the adaptation of the Metaverse, however, malicious users through the gradient updates can steal privacy-sensitive data \cite{xiaolanliu2023}.

Therefore, it is important to emulate the behaviour of an adversary in FL to understand its effects on the learning of the global model and related local personalisation. This will provide insights for designing and testing an appropriate defence for FL. Although accuracy is the most commonly used metric to measure how well an DL model can perform in categorising input samples, it neither measures nor encodes any information about the confidence levels of the underlying model. Concequently, we use uncertainty in order to quantify the impact of the model poisoning attack on the global model.

The existing literature shows that FL is susceptible to poisoning attacks when there are malicious users who can manipulate the data or the model parameters \cite{Aristodemou2023}.Generally, there are three types of adversarial attacks against FL, including model poisoning \cite{Cao_2022_CVPR, baruch2019little, fang2020local}, data poisoning \cite{shi2022, likai2024}, and backdoor attacks \cite{Saha_Subramanya_Pirsiavash_2020, bagdasaryan2019backdoor}. In this paper, we focus only on model poisoning attacks. This is because the other two attacks, data poisoning and backdoor attacks are not sophisticated and they can be detected by a variety of defences and personalisation techniques, such as \cite{li2021ditto}. Specifically, the clients in \cite{li2021ditto} are carefully regularising the weights of the global model into the personalised model. 

Existing model poisoning attack approaches are optimised under a single objective. For example, \cite{baruch2019little} aims to minimise the distance from the original parameters through manipulating the loss function, \cite{fang2020local} computes a vector that changes the direction of the gradients for each model parameter such that the global model will deviate to the wrong direction, \cite{shejwalkar2021manipulating} perturbs the gradients during the weight update to maximise the $L_2$ norm between the benign and malicious gradients. All these works only consider a single optimisation objective and focus on manipulating the gradients. Even though gradients can show the sensitivity between input and output and inform the neurons how well they are performing, they do not consider measuring the uncertainty. Thus, previous studies omit the utilisation of uncertainty as an attack objective and the study of related attack effects on the global and local models in FL unexplored. 

\begin{figure*}[!t]
    \centering
    \includegraphics[width=0.9\linewidth]{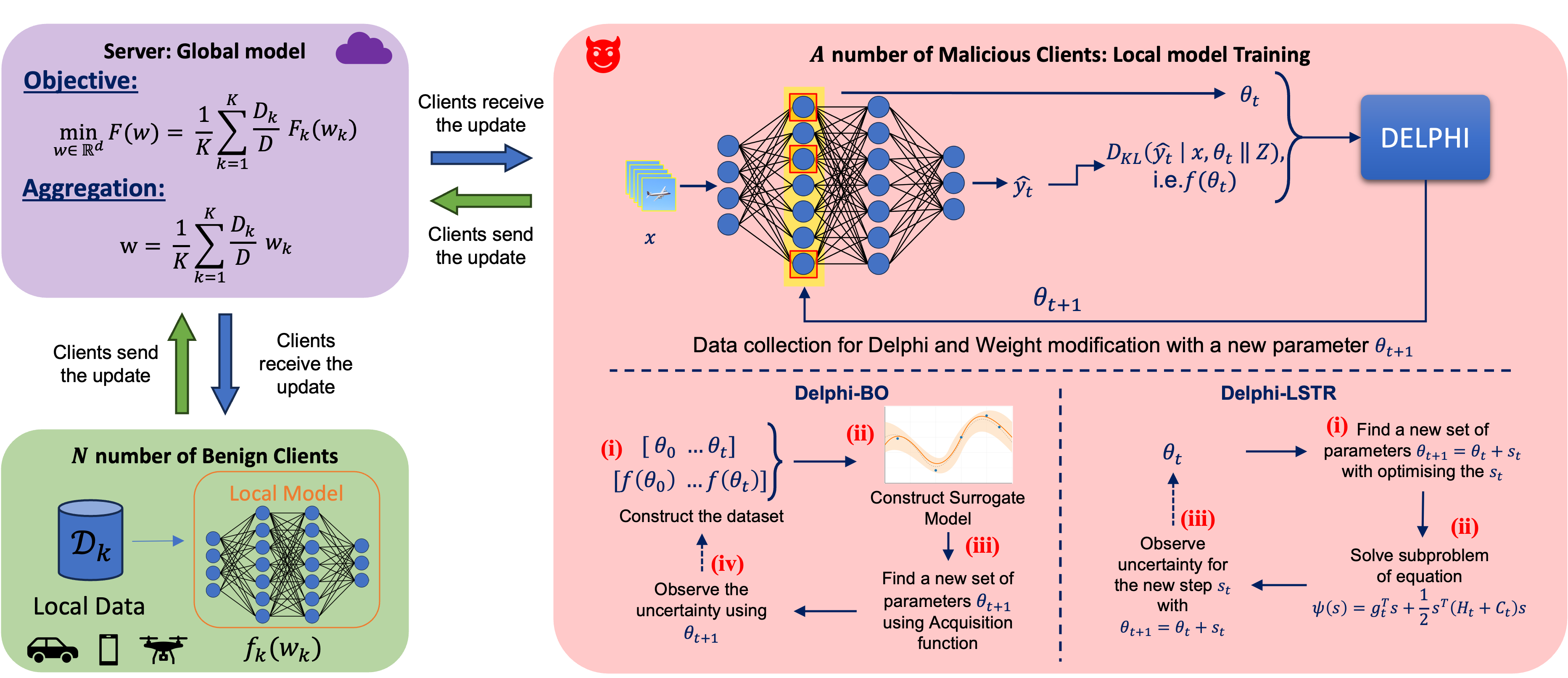}
    \caption{A Federated Learning system, where there is $N$ number of Benign clients and $A$ number of Malicious clients. The Clients are using their own dataset to train their own model. The malicious clients are either train their own model or poison the neurons of the first hidden layer using Delphi. Delphi-BO collects data points to create a surrogate model (i) \& (ii), and finds a new sampling point using an acquisition function (iii). Then modifies with the new parameters $\theta_{t+1}$ the model, and observes the uncertainty (iv). This is repeated for $T$ amount of runs. Delphi-LSTR, search for a new $\theta_{t+1}$ until convergence (i), via solving a subproblem that is to find a small step $s_t$ (ii).Then modifies with the new parameters $\theta_{t+1}$ the model, and observes the uncertainty (iii)}
    \label{fig:basic_diagram}
\end{figure*}

This paper studies the relationship between aforementioned uncertainty and model poisoning attacks in FL and general DL model. We consider that there is a relation that describes the uncertainty induced by the attackers and the modification of the DL model. To achieve this, we propose a strategy named Delphi for model poisoning attacks. Specifically, with Delphi, we search for the optimal model parameters for inducing uncertainty in the DL model using, either white-box optimisation based on Least squares trust region, or black-box optimisation based on Bayesian Optimisation. In addition, we investigate two different ways of modifying the model parameters: 1) keeping a fixed set of neurons per round; or 2) continuously searching for the most significant neurons to generate sufficient attacks to the FL.  As shown in Fig. \ref{fig:basic_diagram}, Delphi identifies the optimal weights of the first hidden layer of the deep neural network using Bayesian Optimisation or Least squares Trust region. Then,  the uncertainty will be induced in the model parameters through the reparametisation of the first hidden layer. Here, instead of manipulating all the parameters in this layer, we manipulate only a small set of neurons. This set of neurons is chosen based on the sensitivity of the neurons on the data input and model's output measured with the $L_2$ norm of the gradient. Similarly, we have chosen the first hidden layer because it has the higher sensitivity between the input and the model's output \cite{Szegedy2014}.
Our numerical results confirm that the attack is more successful when selecting the most significant neurons that will affect the model's prediction in each training round, rather than manipulating the fixed neurons all the time. In addition, we provide a mathematical analysis on measuring the effectiveness and the susceptibility of the model against adversarial attacks.

The main technical contributions are:
\begin{enumerate}
    \item We propose a novel model poisoning mechanism (Delphi), where we evaluate two different types of algorithms including white-box (Least Squares Trust Region - Section \ref{sub:least_square_preliminaries}) and black-box (Bayesian Optimisation - Section \ref{sub:bo_preliminaries}) optimisation to formulate attack strategies. To the best of our knowledge, this is the first work to use both optimisation methods to search for the optimal model parameters for attacks.
    \item With mathematical analysis of attack effectiveness on FedAvg, we provide the foundations for measuring the attack effectiveness (Section \ref{sub:attack_effectiveness}) of model poisoning attacks. We prove that Delphi-BO which chooses the most significant neurons for performing attack in every training round, scores a higher attack effectiveness than manipulating a fixed set of neurons.
    \item Through the comparison between Dephi-LSTR (Section \ref{sub:least_square_preliminaries}) and Delphi-BO (Section \ref{sub:bo_preliminaries}), we demonstrate that the model parameters searched by the BO increases the susceptibility of the model against model poisoning attacks (Section \ref{sub:bo_vs_ls}). Besides, when we are attacking FL with Delphi-BO, the mean predictive confidence is reduced by half.
\end{enumerate}

\section{Related Work}

\subsection{Attacks in Federated Learning}
There is a variety of attacks studied in FL such as data poisoning attacks, backdoor attacks, and model poisoning attacks \cite{Aristodemou2023}.
Data poisoning attacks aim to poison the model through the manipulation of the training data to interfere decision boundary. Methods, such as label-flipping, can mislead the local model \cite{shi2022}, or in \cite{zhang2021} where the authors are using a Generative adversarial network (GAN) to modify the labels of malicious samples. The backdoor attack aims to trigger malicious, incorrect, or unexpected outputs when the specific inputs designed by the attacker are provided to the model. This includes adding a small patch in images with the correct label in order to trigger the backdoor in testing time \cite{Saha_Subramanya_Pirsiavash_2020}. Another example is the training of a similar model to replace the global model through reducing learning rate and the customisation of the loss function in order to deviate the weights through the backpropagation \cite{bagdasaryan2019backdoor}. 

\subsection{Model Poisoning against Federated Learning}
In \cite{likai2024}, the authors created an attack which achieves effectiveness and undetectability by listening to the updates of the local benign and global models and extracting the graph structural correlations between the models and the data features. The authors use an adversarial graph autoencoder to generate a malicious model that maximises the FL training loss. Even though the convergence performance is proved by the authors, such attacks require a high amount of data to get trained and successfully produce a model \cite{likai2024}. 

Another example of model poisoning attack is given with MPAF \cite{Cao_2022_CVPR}. MPAF aims to send local model updates to manipulate the global model in a different direction based on the base model and the global model. The manipulated model is amplified to scale the attack effect. This attack is very similar to \cite{fang2020local}. The authors manipulate the direction of the gradients during the learning process to cause a high error rate during testing. Even though the authors proved their method is transferable to other aggregation functions, this method is computationally expensive. In \cite{baruch2019little}, the authors manipulate the loss function to minimise the distance between the global model and the perturbed model.

Finally, in \cite{shejwalkar2021manipulating}, the noise is added inside the gradients during the weight update such that it can maximise the $L_2$ norm between the benign and malicious gradients. The common ground of all of these attacks is the changing gradients which primarily will change the weights during the network optimisation. Compared to these works, our methods do not consider the manipulation of the gradients but the direct manipulation of weights. In addition, instead of applying a line search approach, we apply a black-box optimisation to search for the model parameters, considering uncertainty as our attack objective.

\begin{table}[!ht]
    \centering
    \caption{Table of Abbreviations}
    \begin{tabular}{|l|l|}
        \hline
        \multicolumn{2}{|c|}{\textbf{General}} \\
        \hhline{==}
        Notation & Definition  \\
        \hline
        $\theta^i$ & parameters of neuron $i$\\
        $\theta$ & parameters of all manipulated neurons\\
        $w_k$ & weight of client $k$\\
        $u(\cdot)$ & uncertainty quantification \\
        $y$ & output of the dataset\\
        $x$ & input of the dataset\\
        $k, K$ & Client\\
        $A$ & Malicious Client \\
        $N$ & Normal Client \\
        $D, D_k, D_a, D_n$ & Overall Dataset\\
        $F_k$ & Objective function\\
        $D_{KL}$ & KL Divergence\\
        $Z$ & Target probability distribution\\
        $f(\cdot)$ & function that describes a relationship \\
        \hhline{==}
        \multicolumn{2}{|c|}{\textbf{Least Square Trust Region}} \\
        \hhline{==}
        Notation & Definition  \\
        \hline
        $l, u$ & lower and upper bound\\
        $f (\cdot)$ & optimisation model\\
        $J_t(s)$ & search space for the algorithm\\
        $\Delta_t$ & ball restrictions\\
        $s_t, S$ & step at time step $t$\\
        $g_t$ & gradient of $f(\cdot)$\\
        $H_t$ & Hessian matrix of $f(\cdot)$\\
        $C_t$ & multiplication of search space $J_t(s)$ and $diag(g_t)$\\
        $\beta_t$ & first order of optimality\\
        $\mu$ & loss tolerance\\
        $\eta$ & accuracy\\
        $\gamma_1$ , $\gamma_2$ & discount factors for $\Delta_k$\\
        \hhline{==}
        \multicolumn{2}{|c|}{\textbf{Bayesian Optimisation}} \\
        \hhline{==}
        Notation & Definition  \\
        \hline
        $\theta_t, f(\theta_t)$ & Candidate $\theta_t$, Observation based on the $\theta_t$\\
        $\Theta$ & \shortstack{Vector including the $f$'s values of all the \\ finite points $[f(\theta_1), \ ..., \ f(\theta_i)]$}\\
        $K(\theta_i, \theta_j)$ & Matern Kernel\\
        $d(\theta_i,\theta_j)$ & Eucledian distance\\
        $K_\nu$ & modified Bessel function\\
        $\rho$, $\nu$ & Matern kernel parameters\\
        $\Gamma(\cdot)$ & Gamma function \\
        $EI(x | \Theta)$ & Expected improvement of $x$ and $\Theta$\\
        $\xi$ & Monte Carlo approximation of $f(x)$\\
        \hhline{==}
        \multicolumn{2}{|c|}{\textbf{Mathematical Analysis}} \\
        \hhline{==}
        Notation & Definition  \\
        \hline
        $w_t, w_t^a, w_t^n$ & \shortstack{weights at time step $t$ of the global model, \\ malicious clients and normal clients respectively} \\
        $\rho$ & attack effectiveness\\
        $\delta$ & malicious client's perturbation in the model\\
        $\epsilon$ &  Expected Perturbation\\
        \hline
    \end{tabular}
    \label{tab:my_label}
\end{table}

\section{Problem Formulation}
In this section, we describe the importance of developing tools and mechanisms to emulate the adversaries' behaviour in the context of distributed learning. Through the proposed methodology, we provide the foundations for proving how the attacker can affect the global model effectively. This is achieved through the manipulation of the first layer of the DL model aiming to induce uncertainty. In order to present our attack, we have to define the threat model and the objective.

\subsection{System Model}
FL is a decentralized deep learning framework where $K$ clients collaborate to train a global model parameterized by $w$ without sharing the local private data $\mathcal{D} = \{\mathcal{D}_1, \mathcal{D}_2, \dots, \mathcal{D}_K \}$, where $\mathcal{D}_k$ represents the local data that a client $k$ holds. To learn the global model parameter $w$, the underlying FL algorithm aims to minimize the following objective function 
\begin{equation} \label{eq:fl_obj}
    \mathop{\min} \limits_{w \in \mathbb{R}^d} F(\bm{w}) = \frac{1}{K} \sum_{k=1}^K \ \frac{D_k}{D}F_k(w_k),
\end{equation}
where $F_k$ is the local training loss adopted by client $k$. In this paper, we consider two representative FL algorithms, namely FedAvg \cite{McMahan2017}, and Krum \cite{Blanchard2017}. Typically, in each communication round, FedAvg updates the global model parameter $w$ via the following model aggregation rule
\begin{equation}\label{eq:general_problem}
w = \frac{1}{K}\sum_{k=1}^{K} \frac{D_k}{D} w_k
\end{equation}
where $w_k$ denotes the local model parameter received from client $k$, which is updated by minimizing the local training loss $F_k$ using a learning algorithm such as stochastic gradient descent (SGD).

\subsection{Threat Model}
In this paper, we poison the local model in order to alter the global models such that to maximise uncertainty. In this scenario, we assume that the attacker is part of the FL process and has all the updates from the global model. In addition, the attacker's manipulated model is in such a way that the server will not detect any anomalies from the attacker's model during early training and the attacker will remain in the training process. We assume that in the adversarial scenario, there are $A$ number of attackers in a configuration of $K$ clients. We can modify eq.(\ref{eq:general_problem}) and rewrite it as 
\begin{equation}\label{eq:general_problem_modified}
    w = \frac{1}{N+A} (\sum^{K-A}_{n=1} w_n + \sum^K_{a=N+1} w_a),
\end{equation}
where $N$ is the number of normal clients and $A$ is the number of malicious or adversarial clients.
Given the above formulation, we can formulate the attackers' objective. The attackers objective is to maximise the uncertainty $u( \; \cdot \;) \; \in \mathbb{R}^d$ in the local model $w_{a}$ to influence the global model and further affect the local updates of the benign clients in the next training rounds. This is given by,
\begin{equation}\label{attack_optimization}
      \theta_a \leftarrow \mathop{\max} 
      u (y, x | \; \bm{w_a})  \; , \; \text{s.t.} \; \{y, \; x\} \in \mathcal{D}_k,
\end{equation}
where $\{y, x\}$ is the training set of the client, and $\theta_a$ is the malicious client's poisoned parameters of the first hidden layers derived using the proposed \textit{Delphi}. Delphi searches for the optimal parameters to modify the layer using optimisation techniques (Bayesian Optimisation and Least Squares Trust Region) and deploys the poisoned parameters  in the first hidden layer of the convolutional neural network (CNN). The use of the first hidden layer is because of the high sensitivity between the input and the output of the neural network. The sensitivity is calculated using the $L_2$ norm of the gradient and we select the top ranked neurons.

\subsection{Delphi's optimisation objective}
The optimisation objective utilises Kullback–Leibler divergence ($D_{KL}$) to measure the statistical distance between the target probability distribution $Z$ and the predicted probability distribution $\hat{y}$ with input data $x$ and parameter $\theta$ for the neuron(s). Therefore, a malicious user aims to minimise the KL divergence in order to obtain the optimal weights. This objective is expressed as follows:
\begin{equation}\label{eq:D_KL_objective}
\begin{aligned}
    \mathop {\min }\limits_{\delta_w} \ D_{KL} (\hat{y} \; | x,\; \theta_i \Vert Z), \; \; \; \; \; \delta_w \leftarrow||w_{t+1} - w_t||_1,
\end{aligned}
\end{equation}
where $\theta_i$ is new weight parameter for the neuron(s) at round $i$, and $\delta_w$ is the constraint of the absolute distance between the previous weights $w_{t}$ and the new weights $w_{t+1}$. In \eqref{eq:D_KL_objective}, the $Z$ is constructed using a discrete probability distribution in which the highest probability for the actual class is set to 0.25, and the probability of the rest of the classes is distributed uniformly.

\section{Methodology}
In this section, we present our methodology for layer optimisation to achieve model poisoning attack against FL. Our methodology called Delphi, utilises two types of optimisation techniques, Least square Trust Region and Bayesian optimisation in order to benchmark which methodology is more efficient and successful. The objective for both optimisation techniques is to obtain optimal layer parameters that can maximise the uncertainty of the model output. We assume that there is a black-box function between the layer parameters of the most significant features of the first layer in the neural network and the uncertainty induced in the model. Since there is no empirical expression between uncertainty and poisoning parameters, we treat the problem in two different contexts. With Bayesian Optimisation technique we treat the uncertainty and model poisoning as a black-box function, and with Least Square Trust Region optimisation, we treat it as a convex optimisation.

Next, the key attack mechanisms of theses two algorithms, and the pipeline of the attack will be presented. We first introduce the preliminaries of the Least Square Trust Region method and the Bayesian optimisation, and show how they could be employed to search for the optimal parameters. Finally, the attack pipeline will be presented

\subsection{Delphi - Least Square Trust Region (Delphi-LSTR) Method} \label{sub:least_square_preliminaries}
Instead of using the intuition of black-box optimisation, and given the fact that the formulated attack problem has a non-linear relation between the input and output, we use Least Square minimisation. For the implementation, we use the function of \verb|scipy.optimise| in order to find the optimal value for the weights. In the following section we explain how the LSTR derives a new parameter $\theta$ for the selected neurons.

\subsubsection{Optimisation Algorithm}

The optimisation of the Least square  in \verb|scipy| uses the trust region reflective methods \cite{least_sqaure_1, least_sqaure_2} to solve a system of equations containing the first-order optimality condition for a bounded minimisation problem as specified below: 
\begin{equation}
    \mathop{\min}_{x \in \mathbb{R}^d} f(\theta) , \; l \leq \theta \leq u,
\end{equation}
where $f( \cdot)$ is the optimisation model which in our case is the quantification of uncertainty based on \ref{eq:D_KL_objective} parameterised with $\theta$, $l$ and $u$ are the lower and upper boundaries, and $\theta$ is the candidate which is the weight parameter we search with Delphi-LSTR.

The new candindate $\theta_{t+1} = \theta_t + s_t$, where $s_t$ is the small step taken by the algorithm at iteration $t$, is being calculated based on the trust region method which typically is ball-bounded and the radius is being controlled by $\Delta$ per iteration $t$. In addition, the algorithm uses the subspace subproblem defined as
\begin{equation} \label{eq:subproblem}
        \mathop{\min}_{s \in \mathbb{R}^d} \psi_t(s) : || J_t s ||_2 \leq \Delta_t, \; 
\end{equation}
where, $s \in S_T$ is the step for the new $\theta_{t+1}$ derived from subspace $S_T \in \mathbb{R}^d$. To solve the minimisation problem of eq. \ref{eq:subproblem}, we need an equation that describes $\psi(s)$, which in this case is a quadratic function. According to \cite{least_sqaure_1, least_sqaure_2}, the $\psi(s)$ is given as follows,
\begin{equation} \label{eq:subproblem_model}
    \psi(s) = g^T_t s + \frac{1}{2} s^T (H_t \ + \ C_t) s,
\end{equation}
where, $g_t$ is the traspose matrix of $\nabla f(\theta_t)$, $H_t$ is the hessian matrix $\nabla ^2 f(\theta_t)$, and $C_t$ is given by $J_t \ diag(g_t) \ J_t^v J_t$. 

The solution of eq. \ref{eq:subproblem} gives us the step $s_t$ which will be added to the $\theta_t$. Note that, at $k=0$ the initial point $\theta_0$ is the initial weight of the neuron. Once the $s_t$ is obtained we need to check whether it satisfies the first order of optimality which is defined as follows,
\begin{equation} \label{eq:ls_rho}
    \beta_t = \frac{f(\theta_t + s_t) - f(\theta_t) + \frac{1}{2}s^T_t C_t s_t}{\psi(s_t)},
\end{equation}
where $r_t$ is the first order of optimality and determines whether the new point moves to the optimal solution. The value of $\beta_t$ must be $>0$, $\beta_t > \mu$ and $\beta_t < \eta$, where $\mu$ is the tolerance for loss, and $\eta$ is the accuracy.

During the iterations until the convergence is achieved, the trust region size $\Delta$ is adjusted by two discount factors $\gamma_1 < 1 < \gamma_2$ based on the range of $\beta_t$. Therefore, the $\Delta_t+1$ as follows,
\begin{itemize}
    \item if $\beta_t \leq \mu$, the $\Delta_{t+1}$ is set in range $(0, \gamma_1 \Delta_t]$
    \item if $\beta_t \in (\mu,\eta)$, the $\Delta_{t+1}$ is set in range $(0, \gamma_1 \Delta_t]$
    \item if $\beta_t \geq \eta$, depends if $\Delta_t$ is less or greater than the lower bound
    \begin{itemize}
        \item if $\Delta_t >$ lower bound, $\Delta_{t+1}$ is set in range $[\gamma_1 \Delta_t, \Delta_t]$ or $[\Delta_t, \gamma_2 \Delta_t]$
        \item if $\Delta_t <$ lower bound, $\Delta_{t+1}$ is set in range $[\Delta_t, min(\gamma_2 \Delta_t,$ upper bound $ )]$ 
    \end{itemize}
\end{itemize}

Further analysis of the Least Square, trust region method can be found in the papers \cite{least_sqaure_1, least_sqaure_2}. The convergence of the algorithms can be achieved on a sufficiently large $T$ number of iterations depending on the size of data to minimise the function \cite{least_sqaure_1, least_sqaure_2}. 

\subsubsection{Attack Pipeline}
Least square Trust Region method treats the optimisation problem as a white-box function where it takes an initial point $\theta_0$ and it is been guided towards the optimality. The attacker with the use of Delphi-LSTR performs the following steps as outlined in \textbf{Algorithm \ref{alg:delphi_ls}}, (i) define the upper and lower bound of the solution, (ii) random initialisation of the perturbation ($s_t$), (iii) iterate until convergence through solving eq. \ref{eq:subproblem_model}, and measuring the convergence with eq. \ref{eq:ls_rho}. 

\begin{algorithm}[ht]
\SetKwData{Left}{left}\SetKwData{This}{this}\SetKwData{Up}{up}
\SetKwFunction{Union}{Union}\SetKwFunction{FindCompress}{FindCompress}
\SetKwInOut{Input}{Input}\SetKwInOut{Output}{Output}
\SetKwComment{Comment}{$\triangleright$\ }{}
\Input{Global model $G(w)$}
\Output{Optimal parameters for neuron}
Receive the global model $G(w)$ \\
Check the loss to be $< 1.5$ \\
Gather the most significant features\\
\text{Define boundaries} \\
\text{Let $s_t$ be the perturbation}\\
\text{Let $f(\theta_t)$ be the uncertainty}\\
\While{$\beta_t < \mu$ and $\beta_t > \eta$}{
    \text{solve} $\psi_t $ \text{on eq.} \ref{eq:subproblem_model} to find $s_t$\\
    \text{calculate} $\beta_t $ \text{based on eq.} \ref{eq:ls_rho}\\
    $\theta_{t+1} = \theta_t + s_t$ \\
}
Send the local model to server $w_a$ 
\caption{Delphi-LSTR - Crafting optimal weights}\label{alg:delphi_ls}
\end{algorithm}

\subsection{Delphi - Bayesian Optimisation (Delphi-BO) Method}\label{sub:bo_preliminaries}
Bayesian Optimisation is a well-studied method for optimising expensive-to-evaluate black-box function \cite{Jones1998}. It leverages a probabilistic surrogate model which describes the hypothesis of the black-box function that we aim to optimise\cite{Snoek2012}. BO has two main aspects that need to be specified before moving to the application of the algorithms. These aspects are, the surrogate model, i.e., a function that describes the hypothesis of the black-box function and it is updated in every iteration with newly observed data, and the acquisition function that relates the belief of the objective function with the input space and aims to find a new sampling point that maximises the objective function \cite{Frazier2018}. Apart from these two aspects, we outline the objective function of the acquisition function to guide the search for new candidates.

\subsubsection{Surrogate Model}
In order to construct a surrogate model, we need to collect $f$'s values at finite points $\theta_1$, ..., $\theta_t \in \mathbb{R}^d$, which can be written into a vector $ \Theta = [ f(\theta_1), ..., f(\theta_t)]$. We suppose each point in this vector is drawn randomly from some prior probability distribution \cite{Frazier2018}. The finite points $\theta_1$, ..., $\theta_t \in \mathbb{R}^d$ are the model parameters for each point $k$ and the $f(\theta_1), ..., f(\theta_t)$ are the values of the uncertainty each time we modify the DL model with parameters $\theta$. We use a normal multivariate Gaussian Process with mean vector $\mu(\Theta)$ and covariance matrix $k(\Theta,\Theta^\prime)$ to represent the surrogate model.
\begin{equation}\label{eq:gpr}
    f \sim GP(\mu(\Theta), k(\Theta,\Theta^\prime)).
\end{equation}

The covariance matrix $K(\Theta,\Theta^\prime)$ is denoted as
\begin{equation}
    K(\Theta,\Theta^\prime) =
  \left[ {\begin{array}{ccc}
    \theta_1, \theta_1 & \cdots & \theta_1, \theta_N\\
    \vdots & \ddots & \vdots\\
    \theta_N, \theta_1 & \cdots & \theta_N, \theta_N\\
  \end{array} } \right]
\end{equation}

The $K(\Theta,\Theta^\prime)$ follows the Matern covariance kernel \cite{pml2Book} which specifies the covariance between two random variables as follows
\begin{equation}
    K(\theta_i, \theta_j) = \sigma^2\frac{2^{1-\nu}}{\Gamma(\nu)}\Bigg(\sqrt{2\nu}\frac{d(\theta_i, \theta_j)}{\rho}\Bigg)^\nu K_\nu\Bigg(\sqrt{2\nu}\frac{d(\theta_i, \theta_j)}{\rho}\Bigg)
\end{equation}
where $d(\theta_i, \theta_j)$ is the Euclidean Distance, $K_\nu$ is a modified Bessel function and $\Gamma(\cdot)$ is the gamma function. The parameter $\nu$ controls the smoothness of the resulting function and is generally set as $\nu = 2.5$, the parameter $\rho$ is a positive parameter that is set to $\rho = 1$. Once the surrogate model is built, we will feed it inside the acquisition function to maxmise or minimise the objective function we have set for our optimisation problem.

Note that, we construct a new surrogate model in every iteration inclusing to the previous observation points, the new observation point from the acquisition function.

\subsubsection{Acquisition Function}
There are multiple acquisition functions, such as Expected Improvement (EI), Probability Improvement, and Upper Confidence Bound (UCB) \cite{Snoek2015}. The most common acquisition function used for BO is expected improvement (EI) \cite{Frazier2018, Zhang2019aml} which aims to maximise the expected improvement over the current best function which describes the observation set. 
\begin{equation}
    EI(x | \Theta) = \mathbb{E}[\max (f(x) - f^*,0)],
\end{equation}
where $f^*$ is the observed best value in $\Theta$ \cite{Frazier2018}, 
and the $EI(x| \Theta)$ is the expected value of the improvement of a chosen $x$ which we have chosen to maximise. We can express the posterior distribution Eq. (\ref{eq:approx}) using Monte Carlo for reducing the computation complexity \cite{Wilson2017, Kingma2013}. 
\begin{gather}
    f(x) = \mathbb{E}[f(\theta) | \xi \sim \mathbb{P}(f(\theta) | \Theta)] \approx \frac{1}{N} \sum_{i=1}^N f(\xi_i),
    \label{eq:approx}
\end{gather}
With this formulation of the posterior distribution, we can define the $qEI$ as in equation (\ref{eq:qEI})
\begin{gather}
    q_{EI}(x) \approx	 \frac{1}{N} \sum_{i=1}^N \max_{j=1,...,q} \{ \max (\xi_{ij} - \textbf{$f$}^*,0)\}, \label{eq:qEI} \\ \xi \sim \mathbb{P}(f(x) | \Theta) \nonumber,
\end{gather}
where $\xi$ is the posterior distribution of the function $f$ at $x$ with the observed data $\Theta$ so far, $N$ Monte Carlo samples and $f^*$ is the observed best function value. 

\begin{algorithm}[t]
\SetKwData{Left}{left}\SetKwData{This}{this}\SetKwData{Up}{up}
\SetKwFunction{Union}{Union}\SetKwFunction{FindCompress}{FindCompress}
\SetKwInOut{Input}{Input}\SetKwInOut{Output}{Output}
\SetKwComment{Comment}{$\triangleright$\ }{}
\Input{Global model $G(w)$}
\Output{Optimal parameters for neuron}
Receive the global model $G(w)$ \\
Check the loss to be $< 1.5$ \\
Gather the most significant features\\
\For{$i\leftarrow 1$ \KwTo $T$ runs}{
    $\hat{y} \leftarrow f_a( \; x\; | \; \theta_i \; )$ \\
    $f(\theta_i) \leftarrow ( D_{KL} (\hat{y} \Vert Z))$ 
    \Comment*[r]{Calculate objectives}
    $\Theta \leftarrow \bigcup_{n = 1}^i \; f(\theta_i) $
    \Comment*[r]{Construct the dataset}
    $f \sim GP(\mu(\Theta), k(\Theta,\Theta^\prime))$
    \Comment*[r]{Construct the surrogate model}
    $\theta_{i+1} \leftarrow \alpha_{qEI} (f)$ 
    \Comment*[r]{Sample a new candindate} 
}
Send the local model to server $w_a$ 
\caption{Delphi-BO - Crafting optimal weights}\label{alg:delphi_bo}
\end{algorithm}

\subsubsection{The Objective Function of Bayesian Optimization }
The process of optimising the acquisition function for searching for new candidates requires an assistance for effective search for the best value due to the high dimensionality. Thus, we observe the uncertainty and the weights of the neurons over each time step. Based on this data, we create Gaussian Process, with an input of the weights and the output of the uncertainty. We use the GP as an objective unction to check whether the output from the posterior of the surrogate model is correct and near the objective of the current weights.

\subsubsection{Attack Pipeline}\label{sub:attack_pipeline}
Assuming that the attacker is participating in part of the training, we apply the poisoning attack to the attacker's local model. Once the attacker receives the global model from the server, the attacker aims to find and manipulate the optimal parameters that will induce uncertainty inside the model using Delphi strategies, including Delphi-LSTR and Delphi-BO. Delphi-BO treats the optimisation problem as a black-box function, which searches the optimal parameters by searching the latent space of the black-box function. The attacker with the use of Delphi-BO performs the following steps as outlined in \textbf{Algorithm \ref{alg:delphi_bo}}: (i) construct a dataset, (ii) initialise a surrogate model, (iii) sample a new set of optimal parameters through optimising acquisition function, (iv) calculate the objective function, (v) expand dataset, and (vi) repeat for $T$ iterations. The optimal solution of the BO can be reached through searching the correct parameters which will describe best the black-box function with a Gaussian Process. Even though, this is an optimisation technique, we aim to create a function that describes the uncertainty inside model parameters rather than converging directly to the optimal solution.

\subsection{Complexity Analysis of Delphi-BO and Delphi-LSTR}
The computational complexity of Delphi variants differs significantly based on their optimisation approaches. Delphi-BO, utilising Gaussian Process (GP) regression, has a time complexity of $O(\lambda\times(n^3 + n^2d + N_sn^2))$ per iteration, where $\lambda$ is the number of neurons, $n$ is the number of observations, $d$ is the input dimension, and $N_s$ is the number of Monte Carlo samples \cite{Frazier2018}. The cubic scaling with observations ($n^3$) arises from GP's covariance matrix inversion, whilst the $n^2d$ term comes from kernel computations. The space complexity is dominated by the GP covariance matrix storage, requiring $O(n^2)$ memory \cite{Wilson2017}. In contrast, Delphi-LSTR employs trust region optimisation with a time complexity per iteration of $O(\lambda\times(nd^2 + d^3))$, where the $d^3$ term is because of the trust region subproblem and the associated matrix operations \cite{least_sqaure_2}. The method's space complexity is $O(d^2)$, primarily for storing the Hessian matrix. This makes LSTR more memory-efficient than BO for problems with many observations but is computationally intensive in high-dimensional spaces. 

\section{Mathematical Analysis of the Attack Effectiveness}
In this section, we propose a measurement of the attack effectiveness in FL. We provide the analysis of the attack effectiveness on FedAvg and that there is an upper bound of the attack. 

\subsection{Definition of Attack Effectiveness}
Let the $c$  be the mean predictive confidence, and $\rho$ be the attack effectiveness. Based on the above notation, we can define the attack effectiveness for FedAvg as the average disruption caused by the malicious users to the global model multiple by the inverse of mean predictive confidence. We can express it as,

 \begin{gather} \label{eq:definition}
    \rho = \frac{1}{c} \cdot \frac{1}{A} \sum^K_{a=N+1} \lVert w_t - w^a_t \rVert^2, \\
    \text{for } c > 0, \ \lvert\lvert w'_{l} - w_l^* \rVert_2 > 0 \nonumber.
 \end{gather}

where, $\frac{1}{A} \sum^K_{a=N+1} \lVert w_t - w^a_t \rVert^2$ is the average disruption caused by the malicious users to the global model. Based on the analysis in the rest of the section, the attack effectiveness has an upper bound depending on the amount of the expected perturbation $\epsilon$ and it is defined as,

\begin{gather} \label{eq:attack_effectiveness}
    \rho \leq \frac{1}{c} \cdot \epsilon^2 (\frac{3 N}{A} + 4).
\end{gather}

\subsection{Bounding Attack}
Before proving that there is an upper bound for the attack when applying Delphi in FL, we first need to define where the malicious users exist inside the general FL problem. First, we recall the equation \ref{eq:general_problem_modified}
\begin{equation}
    w = \frac{1}{N+A} (\sum^{K-A}_{n=1} w_n + \sum^K_{a=N+1} w_a) \nonumber
\end{equation}
To simplify the equation and understand the influence of the optimal weight $w$ and the adversarial contribution $\frac{1}{N+A}\sum^A_{a=0} w_a$, we can rewrite this as
\begin{gather}
    \bm{w}_N = \sum^{K-A}_{n=1} \bm{w}_n \ , \ \ \bm{w}_A = \sum^K_{a=N+1} \bm{w}_a, 
\end{gather}
such that
\begin{gather}
    \bm{w}^* = \frac{1}{N+A} ( \bm{w}_N + \bm{w}_A ).
\end{gather}
Moving forward, we need an expression for $w_A$ and $w_N$. Based on the general problem of the DL and the FL, $w_A$ and $w_N$ can be expressed as follows
\begin{gather}
    w_A = \sum^K_{a=N+1} \frac{D_a}{D} (w - \eta \nabla F_a(w)) + \delta, \\ w_N = \sum^{K-A}_{n=1} \frac{D_n}{D} (w - \eta \nabla F_n(w)),
\end{gather}
where, $\eta$ is the learning rate, $D_a$ and $D_n$, are the local datasets for malicious and normal users respectively, $D$ is the total amount of data available in FL, and $\delta$ is the expected perturbation induced by all the clients. Therefore, we can express the eq. (\ref{eq:general_problem}) in a more analytical form, such that we can have an expression for the expected perturbation inside the system.
\begin{align}
    w = & \frac{1}{K}\sum^K_{k=1} \frac{D_k}{D} w_k, \\
    w = & \frac{1}{N+A} (\sum^K_{a=N+1} \frac{D_a}{D} (w_{t-1} - \eta \nabla F_a(w)) \nonumber \\ 
    & + \sum^K_{a=N+1} \delta \nonumber \\ 
    & + \sum^{K-A}_{n=1} \frac{D_n}{D} (w_{t-1} - \eta \nabla F_n(w))), \\
    \frac{1}{N+A}\sum^K_{a=N+1} \delta & = \ w   \\
    & \ - \frac{1}{N+A} \sum^K_{a=N+1} \frac{D_a}{D} (w_{t-1} - \eta \nabla F_a(w)) \nonumber \\ 
    & - \frac{1}{N+A}\sum^{K-A}_{n=1} \frac{D_n}{D} (w_{t-1} - \eta \nabla F_n(w))) \nonumber
\end{align}
if $\delta = 0$ means that the attack is ineffective. To make it effective, then $\delta$ has to satisfy $\lVert\delta \rVert \geq 0$. This means the $\delta$ should have a lower bound to be effective. Therefore, we define that for an effective function the expectation of $\delta$ should be bounded by $\epsilon$ also referred in this paper as the expected perturbation,
\begin{gather} \label{eq:expected_pert}
    \mathbb{E} [ \   \frac{1}{N+A} \sum^K_{a=N+1} \delta \ ] \geq \epsilon, \ s.t \ \epsilon > 0.
\end{gather}

\subsection{Proof of the Attack Effectiveness} \label{sub:attack_effectiveness}
In this section, we provide the proof for the above attack effectiveness of FedAvg. From the equation (\ref{eq:definition}), we can first analyse the term $\frac{1}{A} \sum^K_{a=N+1} \lVert w_t - w^a_t \rVert^2$, which can be considered as the average discrepancy of the global model from the malicious users
\begin{align} \label{eq:attck_equation_cntd}
    \frac{1}{A} \sum^K_{a=N+1} \lVert w_t - w^a_t \rVert^2  \leq & \ \frac{1}{A} \sum^K_{a=N+1} \lVert \frac{1}{(N+A)} ( \sum^{K-A}_{n=1} \frac{D_n}{D}w^n_t \nonumber \\
    & + \sum^K_{a=N+1} \frac{D_a}{D}w^a_t+\delta) - w^a_t \rVert^2, \\
    = & \ \frac{1}{A} \sum^K_{a=N+1} \lVert \frac{1}{(N+A)} (\sum^{K-A}_{n=1} \frac{D_n}{D}w_t \nonumber \\
    & - \eta \nabla F(w^n_t) \nonumber \\
    & + \sum^K_{a=N+1} \frac{D_a}{D}w_t - \eta \nabla F(w^a_t)) \nonumber \\
    & + \frac{1}{(N+A)}\sum^K_{a=N+1} \delta - w^a_t \rVert^2.
\end{align}
From the eq. \ref{eq:attck_equation_cntd}, it is clear to see that if we exclude the term $\frac{1}{(N+A)}\sum^K_{a=N+1} \delta$, we can consider that the rest of the summation is the sum of the benign model without the adversaries. Therefore, we combine them together to get the average of the weight for a benign model $w^\prime$ plus the perturbation $\delta$ from the malicious users.
\begin{align}
    \frac{1}{A} \sum^K_{a=N+1} \lVert w_t - w^a_t \rVert^2 \leq & \ \frac{1}{A} \sum^K_{a=N+1} \lVert \frac{1}{(N+A)} \sum^{N+A}_{k=1} \frac{D_k}{D}w^k_t \nonumber\\
    & - \eta \nabla F(w^k_t) \nonumber \\ 
    & + \frac{1}{(N+A)} \sum^K_{a=N+1} \delta - w^a_t \rVert^2 \nonumber, \\
    = \ \frac{1}{A} \sum^K_{a=N+1} \lVert w^\prime_t & - w^a_t + \frac{1}{(N+A)} \sum^K_{a=N+1} \delta \rVert^2. \label{eq:converting_benign}
\end{align}
From equation (\ref{eq:converting_benign}), we can consider that all the clients are benign. Based on the definition of $\delta$, and  $E[(X - Y)^2] = E[X^2] -2E[X]E[Y] + E[Y^2]$, we can get an upper bound of the discrepancy.
\begin{align}
    \frac{1}{A} \sum^K_{a=N+1} \lVert w_t - w^a_t \rVert^2 \nonumber \leq & \ \frac{1}{A} \sum^K_{a=N+1} \frac{1}{N+A}(w^\prime_t - w^a_t)^2  \nonumber \\ 
    & + \frac{2}{N + A}(w^\prime_t - w^a_t) \sum^K_{a=N+1}\delta \nonumber \\ 
    & + (\frac{1}{N+A} \sum^K_{a=N+1}\delta)^2 
\end{align}
The term $w^\prime_t - w^a_t$ is clear that it can be subistiuted and considered as the perturbation $\delta$ caused by the $w^a_t$. Then we have,
\begin{align}
    \frac{1}{A} \sum^K_{a=N+1} \lVert w_t - w^a_t \rVert^2 = & \frac{1}{A} \sum^K_{a=N+1} \delta^2 + 2 (w^\prime_t - w^a_t)\sum^K_{a=N+1}\delta \nonumber \\ 
    + &  (\frac{1}{N+A} \sum^K_{a=N+1}\delta)^2 
\end{align}
Based on the property $E[X^2] = E[X]^2$, we substitute the term $(\frac{1}{N+A} \sum^K_{a=N+1}\delta)^2$ with the expected pertrubation $\epsilon$ as defined in the eq. (\ref{eq:expected_pert})
\begin{align}
    \frac{1}{A} \sum^K_{a=N+1} \lVert w_t - w^a_t \rVert^2 = & \frac{1}{A} \sum^K_{a=N+1} \delta^2  \nonumber \\
    & + 2 (w^\prime_t - w^a_t)\sum^K_{a=N+1}\delta + \epsilon^2,\\
    = & \ \frac{1}{A} \sum^K_{a=N+1} \delta^2 + \frac{2 \epsilon}{A} \sum^K_{a=N+1} \delta^2 + \epsilon^2.
\end{align}
We then multiply the first two terms with $\frac{N+A}{N+A}$, to substitute the terms again with the expected perturbation $\epsilon$, we can obtain
\begin{align}
    \frac{1}{A} \sum^K_{a=N+1} \lVert w_t - w^a_t \rVert^2 =  & \ \frac{(N+A)}{A (N+A)} \sum^K_{a=N+1} \delta^2 \nonumber \\ 
    & + \ \frac{2 \epsilon (N+A) \epsilon}{A (N+A)} \sum^K_{a=N+1} \delta + \epsilon^2 \nonumber \\
    = & \ \frac{(N+A)}{A} \epsilon^2 + \frac{2 \epsilon (N+A) }{A} \epsilon + \epsilon^2 \nonumber \\
    = & \ \frac{(N+A)}{A} \epsilon^2 + \frac{2 (N+A) }{A} \epsilon^2 + \epsilon^2 \nonumber \\ 
    = & \frac{3 (N+A) + A}{A}\epsilon^2 \nonumber \\
    = & \frac{3N + 4A}{A}\epsilon^2  \nonumber \\
    \frac{1}{A} \sum^K_{a=N+1} \lVert w_t - w^a_t \rVert^2 \leq & \ \epsilon^2 \ (\frac{3 N}{A} + 4) \label{eq:proof_attack}
\end{align}

Therefore, based on the eq. \ref{eq:proof_attack}, we find out that there is an upper bound for the attack effectiveness. The average amount of disturbance from the attackers in the global model depends on the number of users. This means that less attackers will require more effort to manipulate the users.

\section{Experiments and Simulation Results}
For the experiments\footnote{The implementation of our algorithm can be found in our \href{https://github.com/GrMrWb/Delphi}{Github}, and the codebase is based on \cite{zhang2023pfllib}.}, we evaluate Delphi strategies, i.e., Delphi-BO and Delphi-LSTR, using the CIFAR10 dataset \cite{Krizhevsky2017}, CIFAR100 dataset \cite{Krizhevsky2017}, and compare their attack performances in different adversarial scenarios. The CIFAR10 dataset contains 10 classes of 50,000 images for training and 10,000 images for testing purposes, and CIFAR100 dataset contains 100 classes with 500 images for training and 100 images for testing purposes. The images are RGB with each image size of 32 by 32 pixels.
We distribute the dataset over each user following the distribution of independent and identically distributed (i.i.d.) and imbalanced data. In the imbalanced distribution, the users have all the classes available however the data size is varying with each client. The imbalanced distribution is generated using the Poisson distribution with $\lambda$ set to $0.25\times\text{number of classes}$. This will give us the amount of data that each client will have in percentage. Then, we distribute the data as per the distribution that we have generated. The details are stated in Table \ref{tab:imbalanced_cifar10_cifar100}.

\begin{table}[htbp]
    \centering
    \caption{The Imbalanced data distribution for CIFAR10 and CIFAR100}
    \begin{tabular}{|c|c|c|c|c|}
        \hline
        \multirow{2}{*}{\textbf{Clients}} & \multicolumn{2}{c|}{\textbf{CIFAR10}} & \multicolumn{2}{c|}{\textbf{CIFAR100}} \\
        \cline{2-5}
        & Training & Testing & Training & Testing  \\
        \hhline{=====}
        \textbf{1} & 10000 & 1000 & 9859 & 1000 \\
        \hline
        \textbf{2} & 2500 & 1000 & 6690 & 1000 \\
        \hline
        \textbf{3} & 10000 & 1000 & 7394 & 1000 \\
        \hline
        \textbf{4} & 15000 & 1000 & 9507 & 1000 \\
        \hline
        \textbf{5} & 7500 & 1000 & 8098 & 1000 \\
        \hline
        \textbf{6} & 5000 & 1000 & 8450 & 1000 \\
        \hline
    \end{tabular}
    \label{tab:imbalanced_cifar10_cifar100}
\end{table}

Using the above datasets, we evaluate the attack performance of  Delphi-BO and Delphi-LSTR. The DL model used in our experiments is a CNN with the architecture of AlexNet \cite{Krizhevsky2017}. We optimise the DL model using cross-entropy loss and stochastic gradient decent (SGD). In addition, the Delphi is evaluated against two different FL algorithms, FedAvg \cite{McMahan2017} that is the vanilla FL setting, and the Krum \cite{Blanchard2017} which is a robust and resilient method against the adversaries. The FL is set up with 6 clients from which we are varying the number of attackers from 1 to 3. Also as part of the evaluation, we have tested the algorithm in centralised DL setup. We summarise the experimental setup in Table \ref{tab:experimental_steup}. 

\begin{table}[!ht]
    \centering
    \caption{Summary of the Experimental Setup}
    \centering
    \begin{tabular}{|p{0.35\linewidth} | p{0.35\linewidth}|}
    \hline
    \multicolumn{2}{|c|}{\textbf{Federated Learning Parameters}} \\
    \hhline{==}
    \textbf{Number of Users} & 6 \\
    \hline
    \textbf{Model Aggregation} & FedAvg \cite{McMahan2017} and Krum \cite{Blanchard2017} \\
    \hline
    \textbf{Global \& Local Model} & AlexNet \cite{Krizhevsky2017} \\
    \hline
    \textbf{Loss Function} & Cross Entropy \\
    \hline
    \textbf{Optimiser} & Stochastic Gradient Descent \\
    \hhline{==}
    \multicolumn{2}{|c|}{\textbf{DL model: AlexNet}} \\
    \hhline{==}
    \textbf{No. of Parameters} & 60 million\\
    \hline
    \textbf{Layers} & 8\\
    \hline
    \textbf{Kernel Size} & (3 $\times$ 3), (3 $\times$ 3), (3 $\times$ 3), (3 $\times$ 3), (3 $\times$ 3) \\
    \hhline{==}
    \multicolumn{2}{|c|}{\textbf{Dataset}} \\
    \hhline{==}
    \textbf{Dataset} & CIFAR10 \cite{Krizhevsky2017}, CIFAR100 \cite{Krizhevsky2017}, 
    \\ 
    \hhline{==}
    \multicolumn{2}{|c|}{\textbf{Experiments}} \\
    \hhline{==}
    \textbf{Adversaries} & varying from 1 to 3 \\
    \hline
    \textbf{Normal Users} & varying from 3 to 5  \\
    \hline
    \textbf{Attacks} & Delphi(LS), Delphi(BO)  \\
    \hline
    \textbf{Selecting Neurons} & Fixed and Dynamic set of Neurons \\
    \hline
    \end{tabular}
    \label{tab:experimental_steup}
\end{table}

In the following sections, we discuss 5 different studies to analyse the performance of the Delphi attack strategies and attack effectiveness in different scenarios. We measure the attack performance using accuracy, mean predictive confidence and entropy. Note that, we have two types of selection for the set of neurons, fixed scheme and dynamic searching (DS) a set of neurons, meaning that in the fixed scheme, we manipulate the same set of neurons in every iteration, and in DS scheme, we select a new set of neurons in every iteration based on the sensitivity of the neurons on the data input and model's output, measured with the $L_2$ norm of the gradient.

\subsection{Comparing the Attack Performance of Bayesian Optimisation and Least Squares} \label{sub:bo_vs_ls}
\begin{figure}[tb]
\centering
    \centering
    \subfloat[\label{fig:LS_Vs_BO_Neurons_cifar10_imbalanced} CIFAR10]{%
        \includegraphics[width=\linewidth]{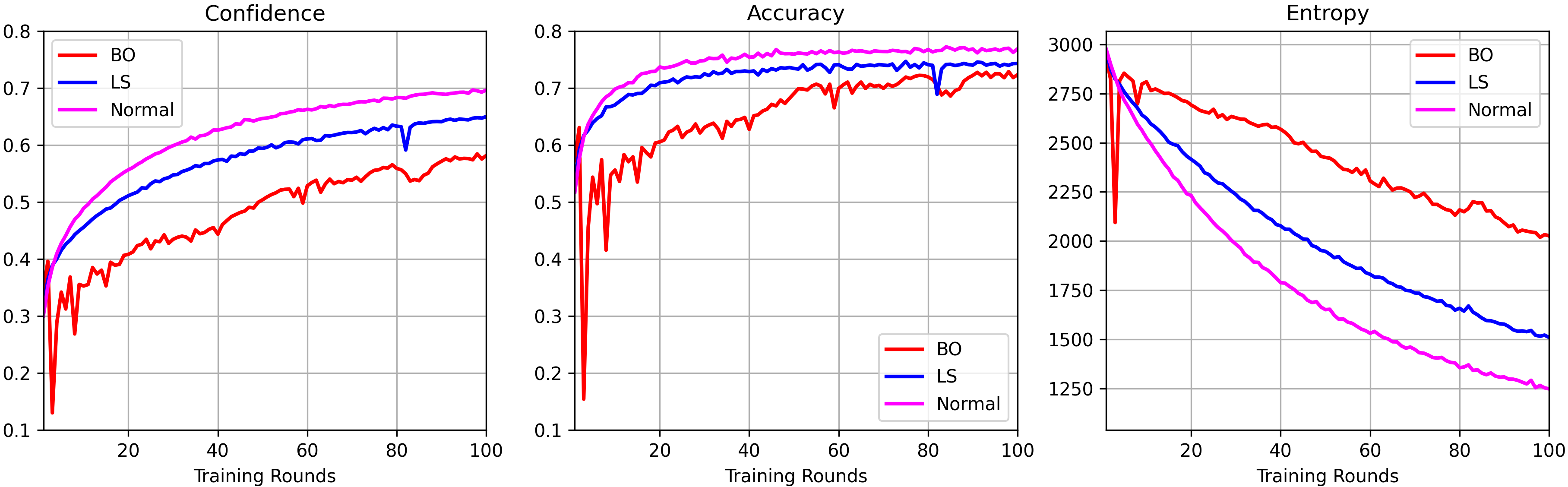}}
    \hfill\\
    \subfloat[\label{fig:LS_Vs_BO_Neurons_cifar100_imbalanced} CIFAR100]{%
        \includegraphics[width=\linewidth]{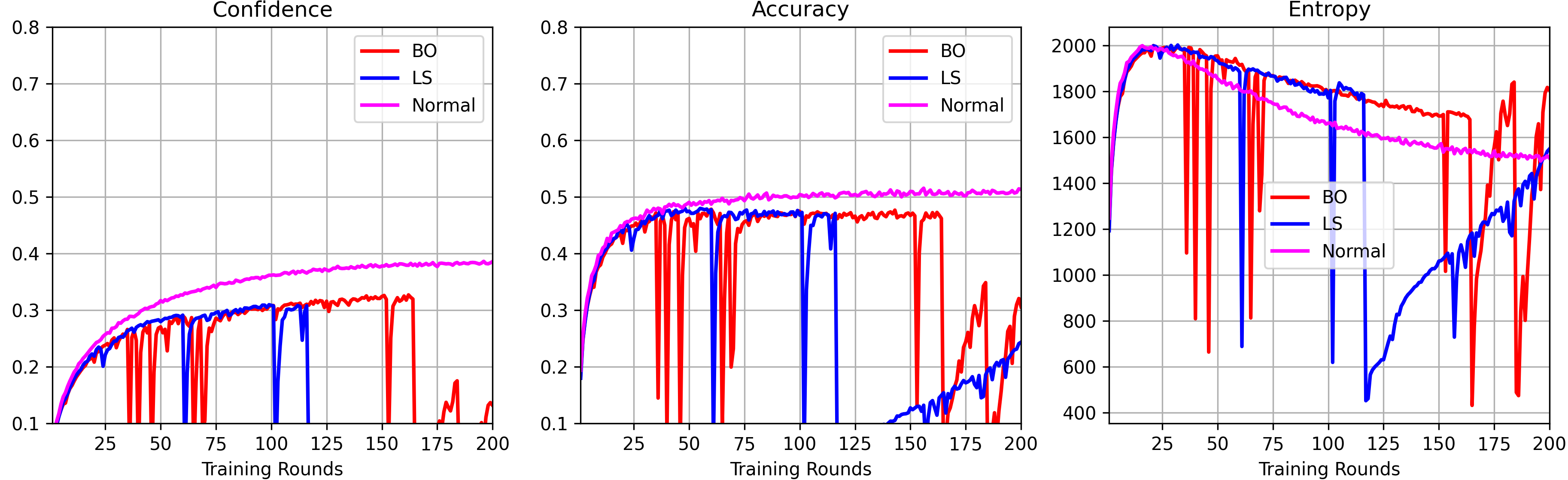}}
    \hfill
\caption{Comparison between Bayesian Optimisation and Least Squares with fixed neuron selection scheme and datasets of CIFAR10 and CIFAR100. We show  the effects of mean predictive confidence, accuracy and entropy when attacking 5 neurons.}
\label{fig:ls_vs_bo_imbalanced_dataset}
\end{figure}

In this study, we compare the two optimisation techniques used for developing the Delphi-BO and Delphi-LSTR. We demonstrate the capabilities of Least square and Bayesian Optimisation with imbalanced data distribution among the users. 
To recap, the difference between LSTR and BO is that LSTR considers the relation between uncertainty and optimal weight parameter a white-box function and BO as a black box function.
In Fig. \ref{fig:LS_Vs_BO_Neurons_cifar10_imbalanced} and Fig. \ref{fig:LS_Vs_BO_Neurons_cifar100_imbalanced}, we illustrate the predictive confidence of the global model with 2 malicious users, and we are manipulating 5 neurons in the first hidden layer. The CIFAR10 and CIFAR100 dataset is used in this experiment. Also, note that all the results are generated under the fixed neuron selection scheme. 

It is apparent that the BO has better performance than the LSTR in both datasets. There is higher level of uncertainty when we are looking into entropy and mean predictive confidence. Even though BO is better, we need to verify this with more experiments which are demonstrated on the next sections.

Apart from the FL algorithm, in Table \ref{tab:centralised}, we demonstrate the attack performance of the proposed Delphi strategies in a centralised DL algorithm trained with CIFAR10 and CIFAR100 dataset. From Table \ref{tab:centralised}, it is apparent that the Delphi-BO algorithm is more successful than the Delphi-LSTR. It has scored higher than the Delphi-LSTR under both accuracy and mean predictive confidence. 

\begin{table}[!h]
    \centering
    \caption{Centralised DL - Model trained for 50 epochs}
    \begin{tabular}{|c|c|c|c|c|}
        \hline
        \multirow{2}{*}{\textbf{Model}} & \multicolumn{2}{c|}{\textbf{CIFAR10}} & \multicolumn{2}{c|}{\textbf{CIFAR100}} \\
        \cline{2-5}
         & \textbf{Accuracy} & \textbf{Confidence} & \textbf{Accuracy} & \textbf{Confidence}\\
        \hhline{=====}
        Normal & 0.7677 & 0.7022 & 0.5177 & 0.3837 \\
        \hline
        Delphi(LS) & 0.6946 & 0.6243 & 0.4026 & 0.2658 \\
        \hline
        Delphi(BO) & \textbf{0.2442} & \textbf{0.2246} & \textbf{0.1411} & \textbf{0.1253}\\
        \hline
    \end{tabular}
    \label{tab:centralised}
\end{table}

\subsection{Varying the amount of Neurons}
In this section, we are varying the amount of neurons and the malicious users inside the FL. We have chosen to manipulate 5 or 10 neurons and vary the number of malicious  from 1 to 3. All the experiments were conducted in imbalanced data distribution of CIFAR10 and we measure the mean predictive confidence through out the training. The manipulated neurons are being chosen in the beginning of the training and they remain fixed through out the training process.

\begin{figure}[tb]
\centering
    \centering
    \subfloat[\label{fig:LS_Vs_BO_Neurons_5_imbalanced} 5 Neurons]{%
        \includegraphics[width=\linewidth]{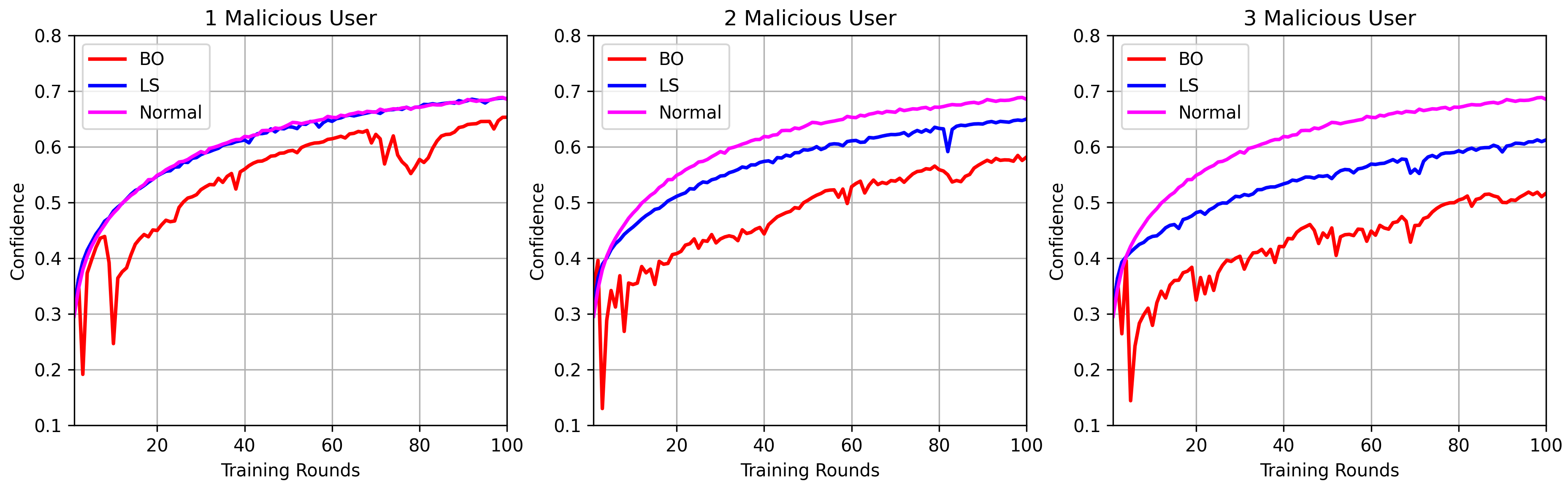}}
    \hfill\\
    \subfloat[\label{fig:LS_Vs_BO_Neurons_10_imbalanced} 10 Neurons]{%
        \includegraphics[width=\linewidth]{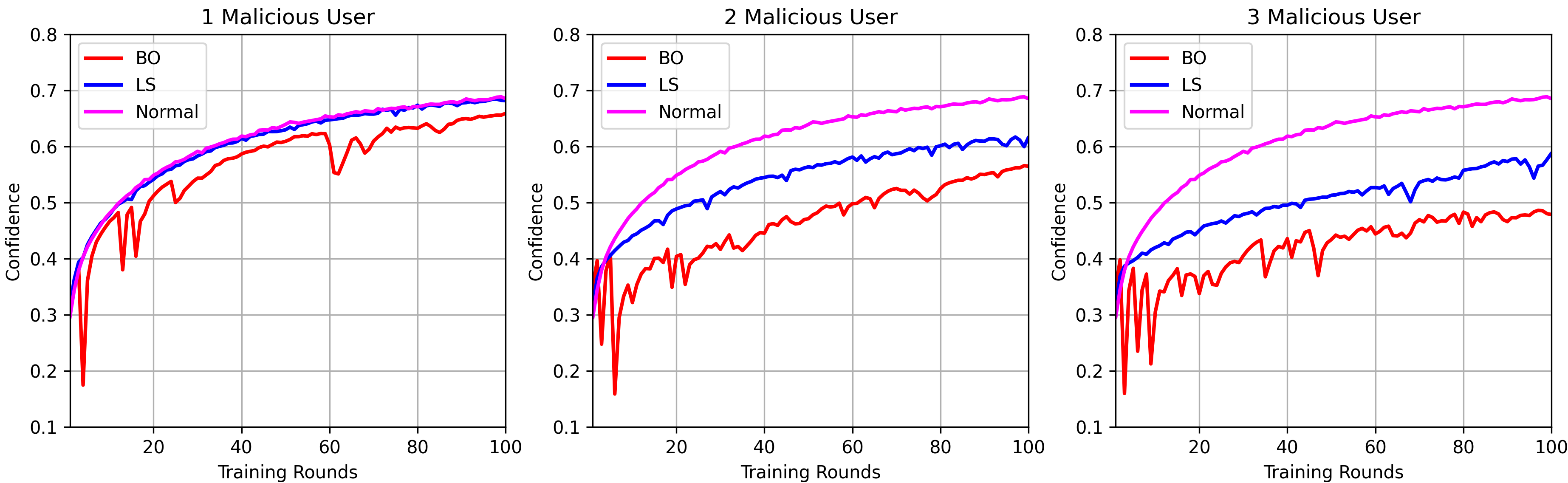}}
    \hfill
\caption{Comparison between Bayesian Optimisation and Least Squares, with varying number of malicious users and neurons  with fixed neuron selection scheme. We show  the effects of mean predictive confidence.}
\label{fig:ls_vs_bo_imbalanced}
\end{figure}

From Fig. \ref{fig:ls_vs_bo_imbalanced}, it is clear that the Delphi-BO causes more disturbance to the FL comparing to the model without any malicious users and Delphi-LSTR. Particularly, the Delphi-LSTR is effective when more than one malicious user exists in the FL network, and when we manipulate more neurons. This shows that Delphi-LSTR does not converge towards the optimal parameters, probably because it requires more freedom with respect to the model's parameters that are required to get manipulated for sufficient model poisoning. 

Taking into the account the findings in the subsection \ref{sub:bo_vs_ls} and from this section, Delphi-BO is more capable than Delphi-LSTR, for the amount of information that the attacker is able to manipulate in the model. Based on this, for the rest of the studies, we will use Delphi-BO for our comparisons and benchmarks. 

\subsection{Dynamic Searching for Neurons} \label{sub:dynamic_search}
Choosing the neurons effectively is very important to poison the model correctly. Therefore, to test how effectively the poison is being applied, we consider two different approaches, which are fixed neurons selection scheme and the DS scheme. 
We have tested this hypothesis in IID and Imbalanced data among the users using CIFAR10 dataset, and we obtain the results under the mean predictive confidence in the global model, the accuracy, and the entropy. We use Delphi-BO to craft the model parameters for the layer. Again, in this experiment there are two malicious users.

\begin{figure}[h]
    \centering
    \includegraphics[width=\linewidth]{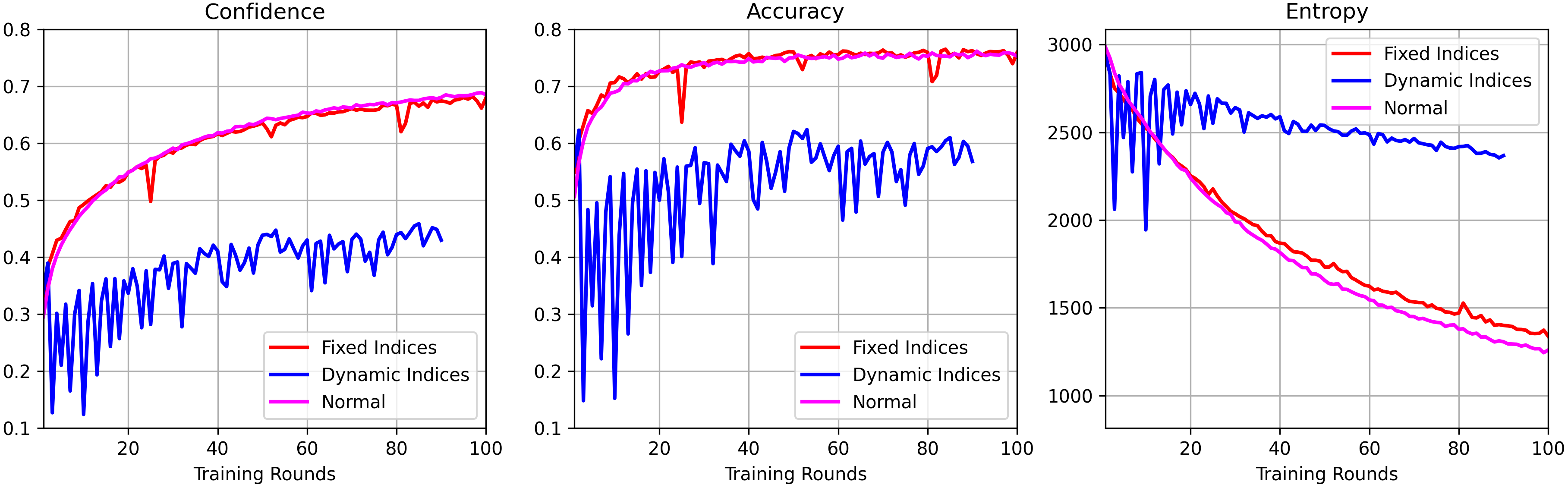}
    \caption{Comparison of manipulating with fixed and dynamic set of neurons, in the setting of IID data distribution}
    \label{fig:fixed_indices_vs_continues_iid}
\end{figure}

Beginning with IID, in Fig. \ref{fig:fixed_indices_vs_continues_iid}, it is clear that the DS neuron selection scheme can lead to lower predictive confidence. It is apparent that under all the three performance metrics,  the DS scheme is showing attack effectiveness. In addition, we can see that with the fixed scheme, there is no difference between the global model with benign models and adversarial models. This phenomenon probably is because the chosen neurons are not very important to the model and they are being discarded from the fully connected layers during training. 

\begin{figure}[h]
    \centering
    \includegraphics[width=\linewidth]{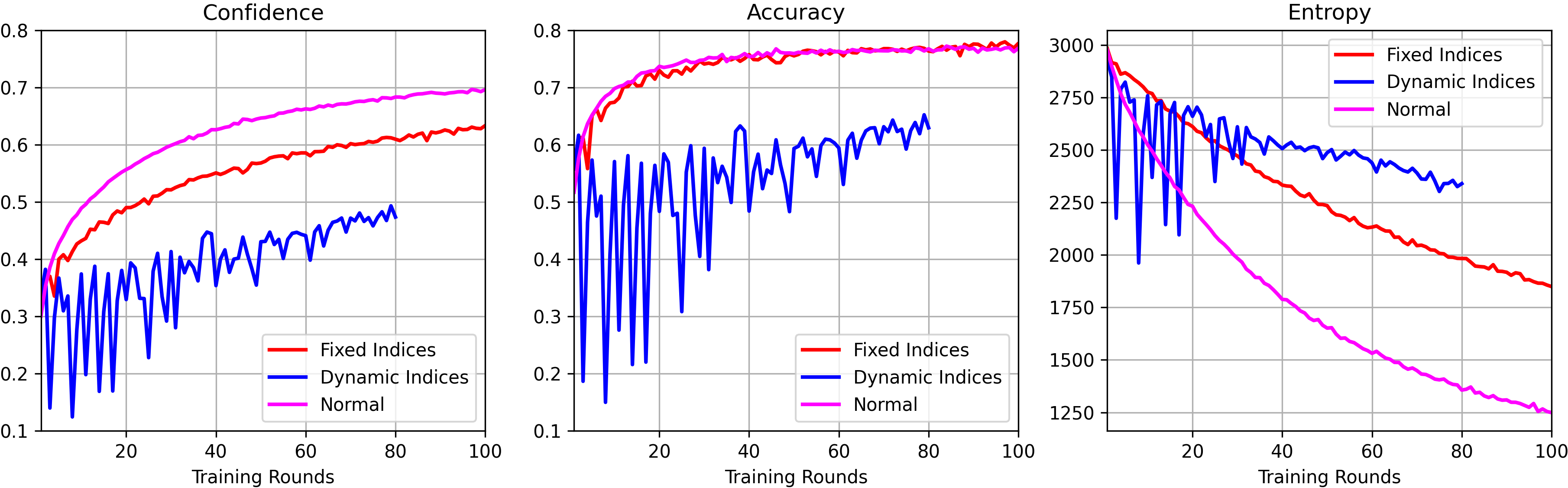}
    \caption{Comparison of manipulating a fixed or dynamic set of neurons, in the setting of Imbalanced data distribution}
    \label{fig:fixed_indices_vs_continues_imbalanced}
\end{figure}

While testing this hypothesis with Imbalanced data, it allows us to check whether the global model will be affected by the poisoning attack when the attacker has more data available for each class than the rest of the benign clients. This helps us to understand how data distribution can affect the global model and the selection of the neurons. In Fig. \ref{fig:fixed_indices_vs_continues_imbalanced}, we use the same settings as in Fig. \ref{fig:fixed_indices_vs_continues_iid} with the key difference in data distribution among the users. Different from the IID scenario, we can see that the fixed neuron selection scheme has more impact on the global model than the DS neuron selection scheme. This shows that data distribution does have an impact on the learning ability of the model.  We notice that the DS scheme also causes a small reduction in model's accuracy. Particularly, the mean predction confidence interval is increased slightly, as well as the accuracy and the entropy. 

\subsection{Measuring Attack Effectiveness}
\begin{figure}[h]
    \centering
    \includegraphics[width=0.6\linewidth]{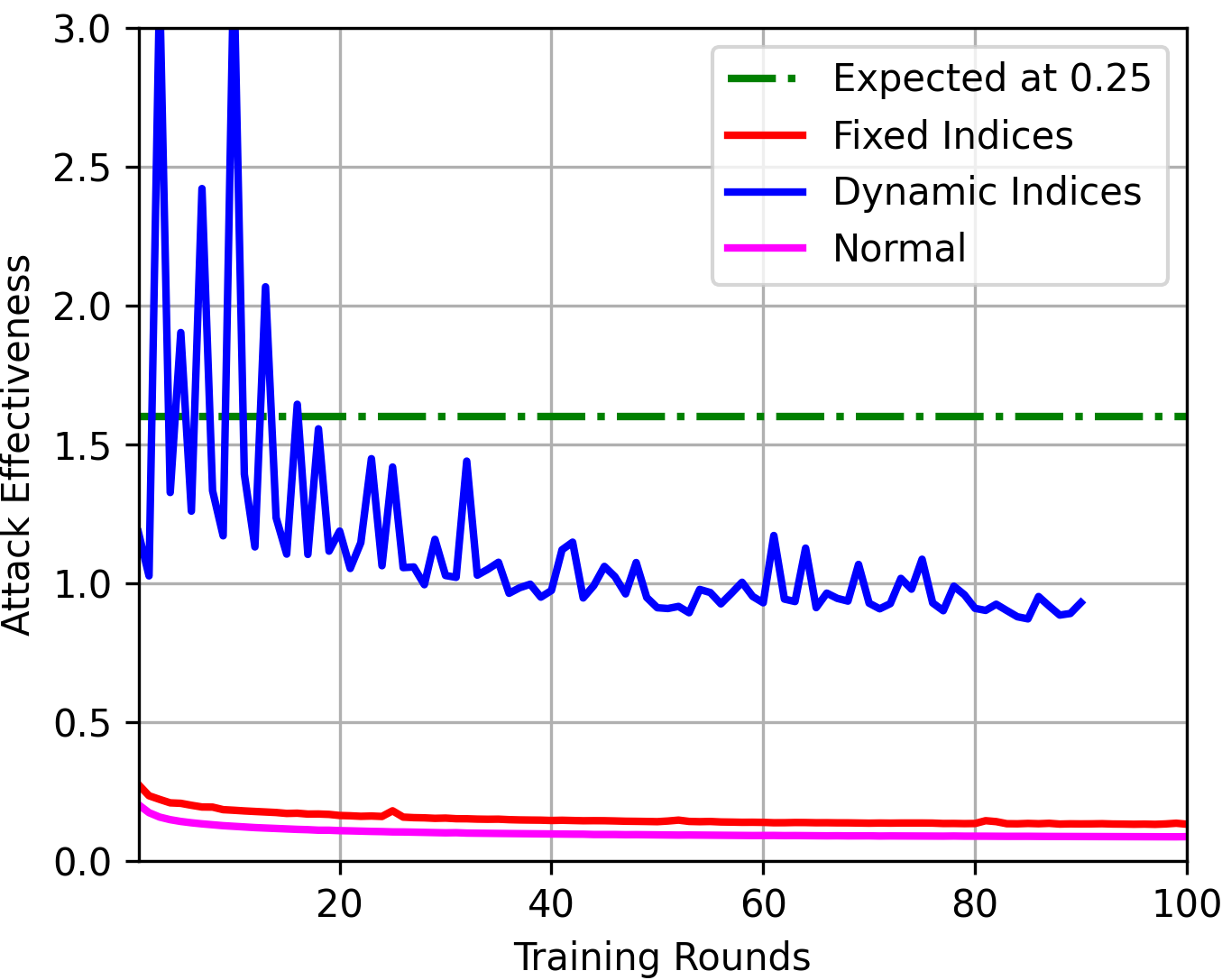}
    \caption{Attack effectiveness for Delphi-BO in the context of fixed neurons or continuously changed neurons for manipulation in IID scenario}
    \label{fig:attack_effectiveness_iid}
\end{figure}

\begin{figure}[h]
    \centering
    \includegraphics[width=0.6\linewidth]{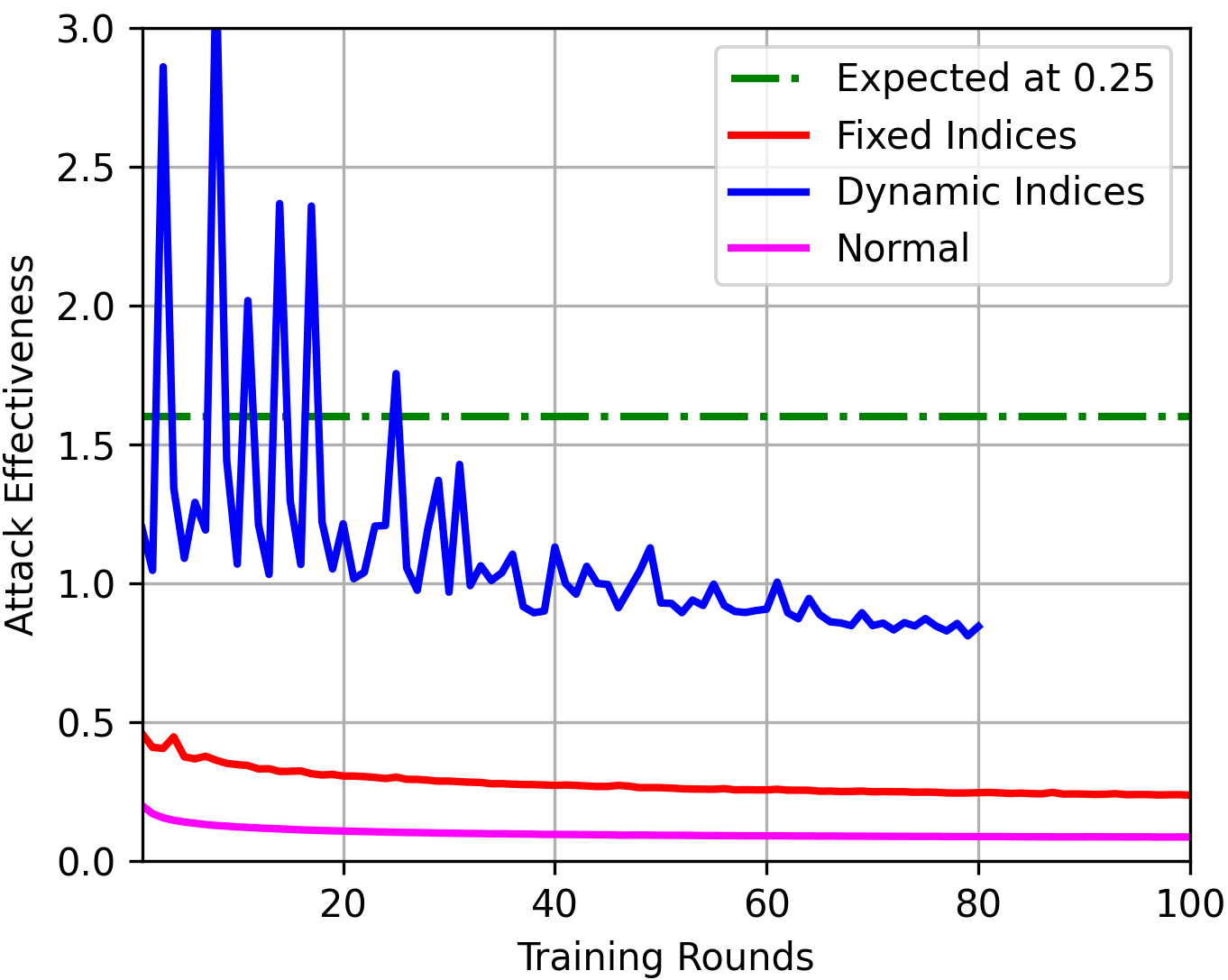}
    \caption{Attack effectiveness for Delphi-BO in the context of fixed neurons or continuously changed neurons for manipulation in Imbalanced scenario}
    \label{fig:attack_effectiveness_imbalanced}
\end{figure}

In this study, we are testing the attack effectiveness defined as in equation (\ref{eq:attack_effectiveness}). The experimental setting is following the same settings as the section \ref{sub:dynamic_search}. We have 2 attackers and 4 benign users in the FL network, and use imbalanced and IID data distributions with CIFAR10 dataset. In the equation (\ref{eq:attack_effectiveness}), there are two terms that we can consider as constant throughout the process. This is the term $\frac{3 N}{A} + 4$ which is equal to $10$, and the second term is the expected perturbation which is $\epsilon = 0.2$ and $\epsilon^2 = 0.04$ for DS neuron selection and $\epsilon = 0.1$ and $\epsilon^2 = 0.01$ for fixed neuron selection in IID and $\epsilon = 0.125$ and $\epsilon^2 = 0.015$ in Imbalanced data distribution. Also, in Fig. \ref{fig:attack_effectiveness_iid} and Fig. \ref{fig:attack_effectiveness_imbalanced}, we have plotted the expected upper bound if the objective function has been minimised completely with the predictive confidence at 0.25. Anything above that line, the attack probably is going to be detected from a defensive mechanism.

In Fig. \ref{fig:attack_effectiveness_iid} and Fig. \ref{fig:attack_effectiveness_imbalanced}, we have plotted the equation (\ref{eq:attack_effectiveness}) over the training rounds. From both figures, when we use the DS neuron selection scheme, we can extract that in the early stage of training, the attack effectiveness is higher than at the stages where the model has learned the representations. In addition, we can see that the attack effectiveness in both IID and imbalanced data distribution remains almost the same. Apart from the fact that the DS scheme is more effective, we should highlight, that in the early stage the attack exceeds its effectiveness, resulting to be above the expected behaviour. Thus, in case there is a defensive mechanism, the malicious users may be detectable in the early stages, rather than in the later stages.

An important thing that we need to highlight is the behaviour of a global model without attackers. With the current formulation of the equation (\ref{eq:attack_effectiveness}), we should consider that there is at least one attacker in FL. In our case, we consider the same number of benign and malicious users as the rest of the cases. However, we consider that the expected perturbation is very small. That is why the attack effectiveness is very low. 

\subsection{Defending against Model Poisoning}
Another hypothesis that we need to test is  how much resiliency the defending mechanisms in FL can have when they are dealing with model poisoning attack. Thus, we are testing the Delphi-BO in the scenario where the FL system run the Krum \cite{Blanchard2017} aggregation. Krum \cite{Blanchard2017} is one of the aggregation used to benchmark the susceptibility of FL in the presence of an attack. It measures the distance between the weights from each client, and chooses clients to aggregate that are closer together, meaning that a malicious client can be excluded by the aggregation. In this ablation study, we use the same settings as above where we have 2 attackers and 4 beingn users, tested in IID and imbalanced datasets. The attacker can manipulate 5 or 10 neurons. 

\begin{figure}[h]
    \centering
    \includegraphics[width=\linewidth]{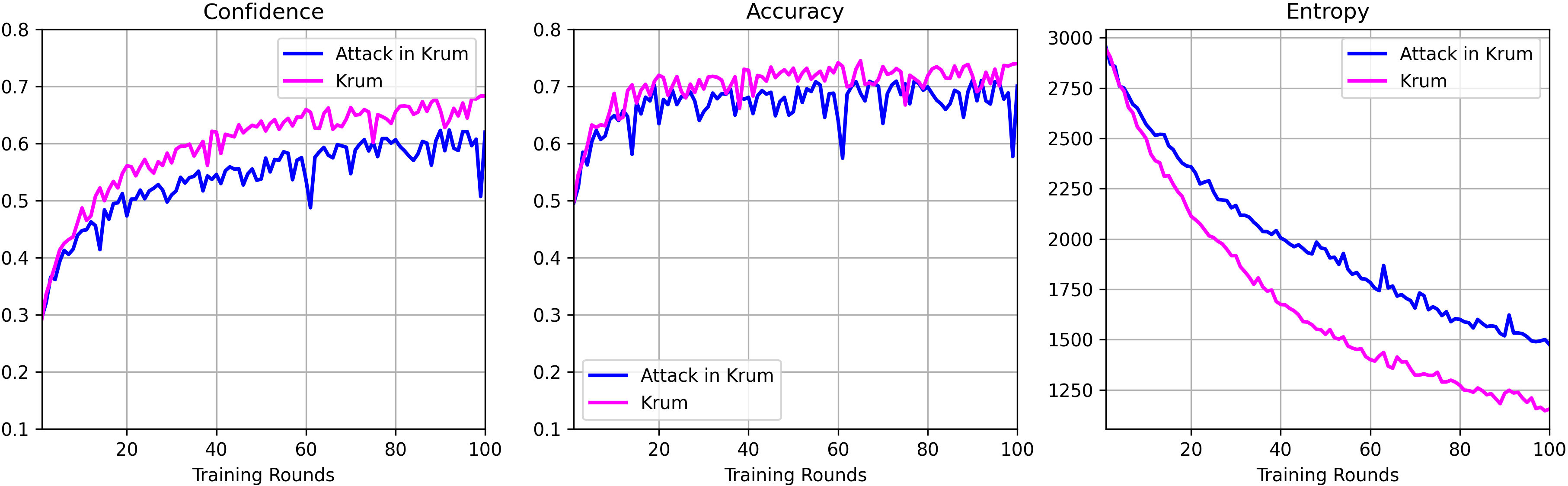}
    \caption{Comparison of Krum in IID data distribution}
    \label{fig:Krum_iid}
\end{figure}

\begin{figure}[h]
    \centering
    \includegraphics[width=\linewidth]{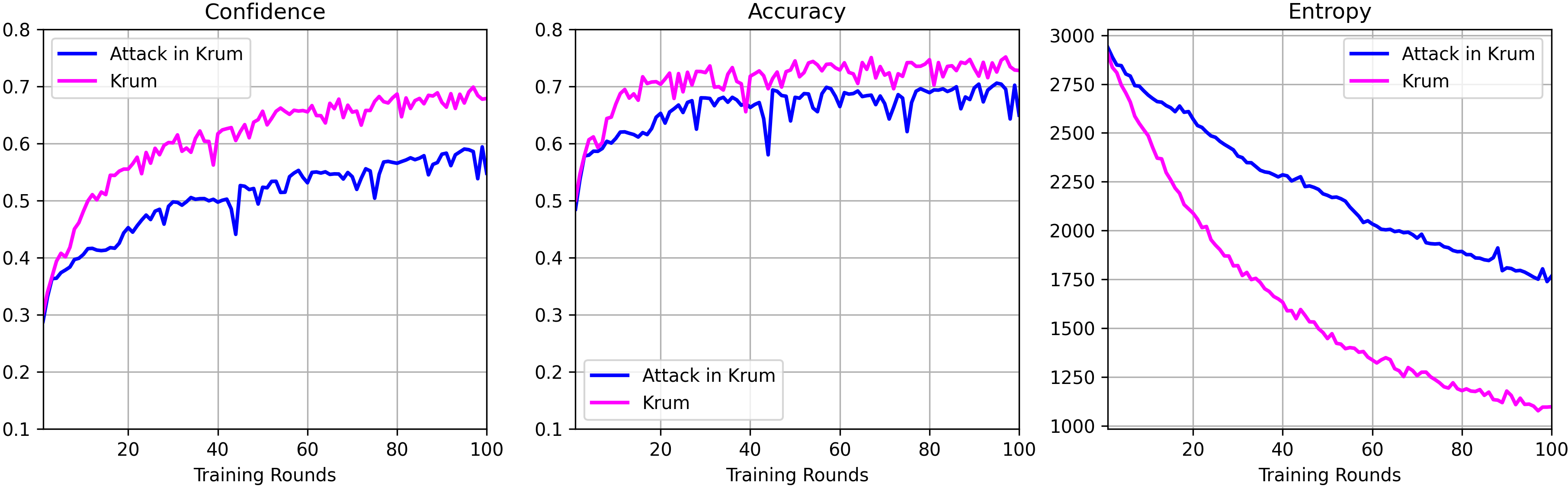}
    \caption{Comparison of Krum in Imbalanced data distribution}
    \label{fig:Krum_imbalanced}
\end{figure}

In  Fig. \ref{fig:Krum_iid} and \ref{fig:Krum_imbalanced}, we can see that there is no difference between the IID settings and the Imbalanced data distribution. This is because the weights are more sparse in the Imbalanced than IID scenario, so that the attacker's weight vector is closer to the weights of the rest of the benign users. Compared to the FedAvg, the performance of the global model has slightly higher predictive confidence in Krum.

Therefore, the attack needs to be improved through the expansion of the attack in the rest of the layers or towards the rest of the neurons of the manipulated layer.

\section{Conclusion}
We have proposed an attack strategy based on two optimization strategies, Bayesian Optimization (BO) and Least Square Trust Region (LSTR), named as Delphi-BO and Delphi-LSTR, to evaluate the uncertainty of the global model output by inducing the malicious weight parameters in the first layer of the local model in federated learning (FL). We also provided a mathematical analysis for attack effectiveness, which shows that the amount of disturbance from the attackers in the global model depends on the number of global users, and the effort required by the attackers will be higher.
Numerical results demonstrated that the Delphi-BO using a black-box optimisation is superior to the Delphi-LSTR using the white-box optimisation. In addition, the dynamically searched neurons for attacking the global model in each training round  is more efficient rather than attacking fixed neurons every round. Apart from this,by presenting that the Delphi attack strategies are effective in different datasets and FL algorithms, such as in Krum, we observed that the attack without exercising the full power of Delphi can penetrate the aggregation.

The attack effectiveness mathematical analysis is based on the mathematics of FedAvg. In our future work, we will expand the attack effectiveness and derive mathematical proofs for different types of aggregation functions and FL schemes. In addition,  a more comprehensive analysis for a variety of DL models, such as transformers will be provided.

\bibliographystyle{IEEEtran}
\bibliography{MasterBiblio}

\begin{IEEEbiography}
[
{\includegraphics[width=1in,height=1.25in,clip,keepaspectratio]{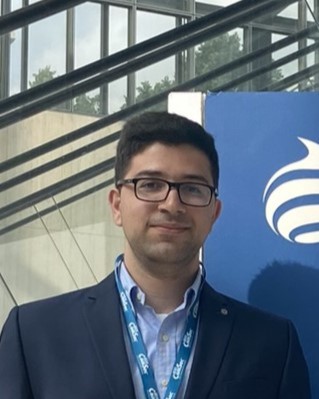}}
]{Marios Aristodemou}
received the MEng Systems Engineering degree from Loughborough University, U.K in 2021. 
He has finished his Ph.D. degree with the Wolfson School of Mechanical, Electrical and Manufacturing Engineering, Loughborough University, U.K. He is currently a Post Doctoral Research Associate in University of York. He has being involved in various projects in Wireless Communications and Cyber Security, applying distributed learning and reinformcenet learning.
His research interests include the Artificial Intelligence and Metaverse, including topics such as the security and trustworthiness, personalised distributed learning, privacy preserving for distributed learning and lifelong learning. 

\end{IEEEbiography}

\begin{IEEEbiography}[
{\includegraphics[width=1.2in,height=1.25in,clip,keepaspectratio]{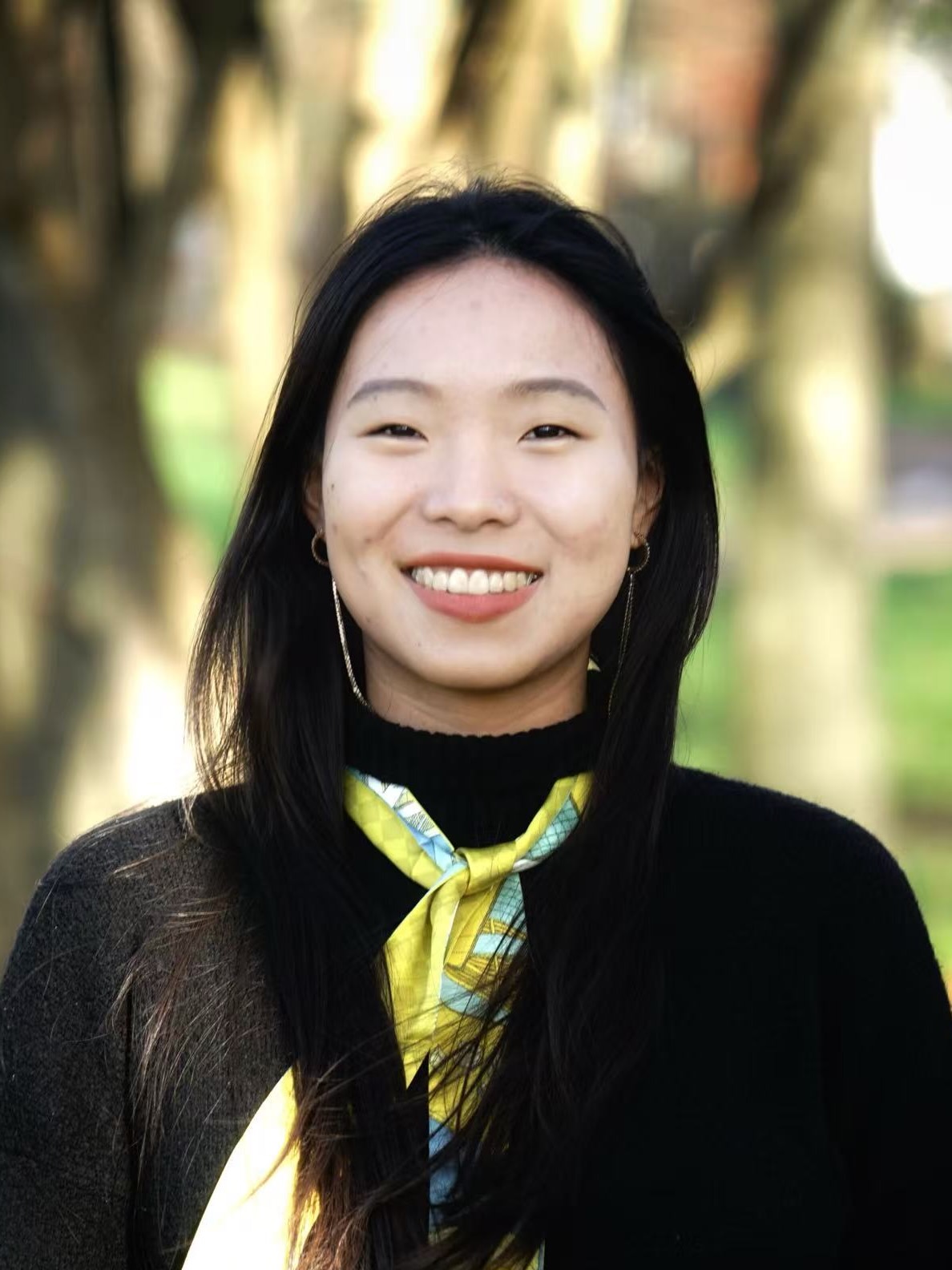}}
]{Xiaolan Liu }
received the Ph.D. degree from Queen Mary University of London, U.K. in 2021. She was a Research Associate at King’s College London from 2020-2021. She was a Lecturer at Loughborough
University London from 2021-2024. Since 2024, she has been a Lecturer with School of Electrical, Electronic and Mechanical Engineering at University of Bristol, U.K. Her current research interests include wireless distributed learning, multi-agent reinforcement learning for edge computing, and machine learning for wireless communication optimization. 
\end{IEEEbiography}

\begin{IEEEbiography}[
{\includegraphics[width=1in,height=1.25in,clip,keepaspectratio]{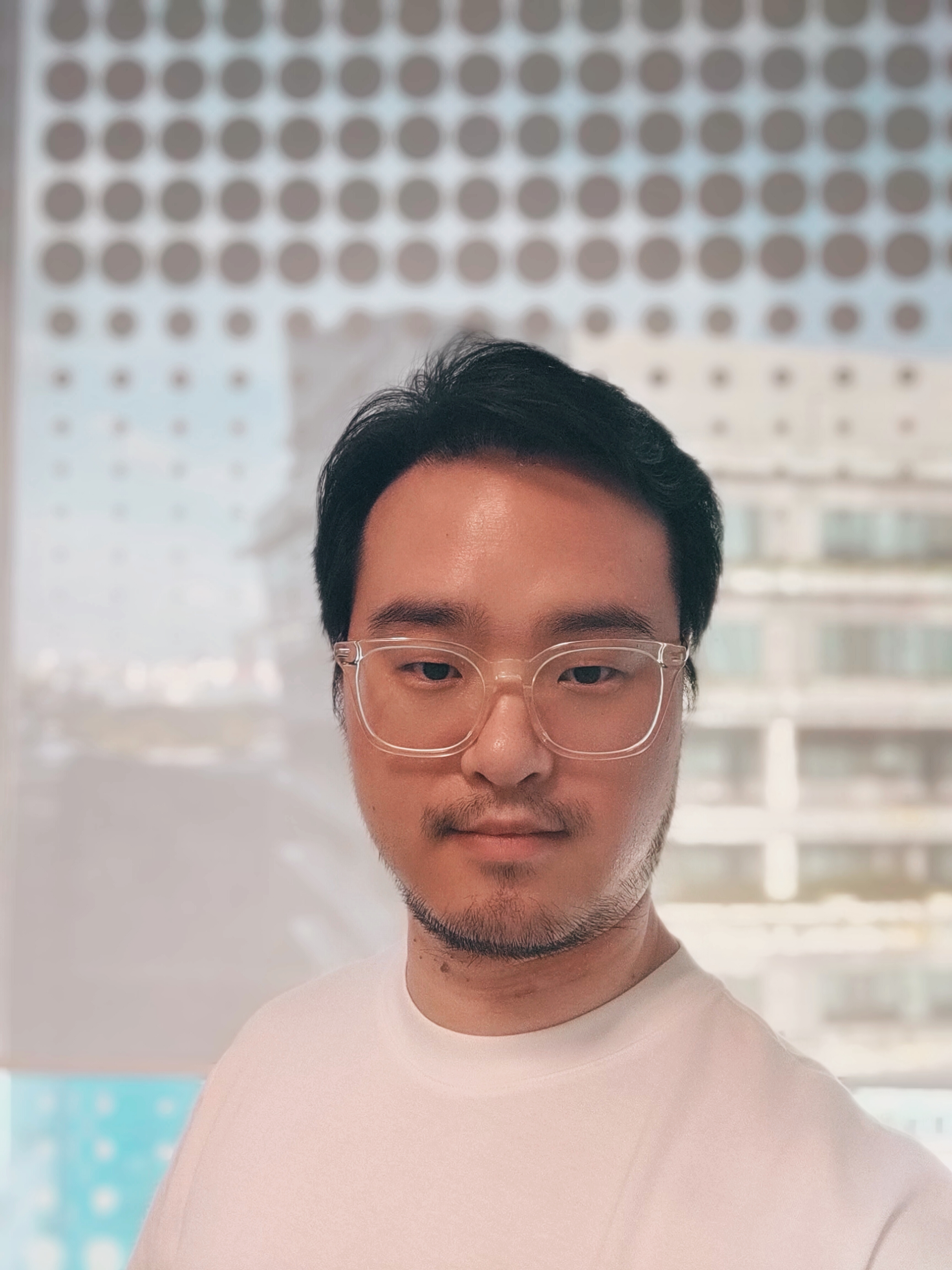}}
]{Yuan Wang}
received his Ph.D. degree from Nanyang Technological University, Singapore. He is now a senior scientist with the department of Computing and Intelligence, Institute of High Performance Computing, under A*STAR, Singapore. His current research mainly focus on decentralized and trustworthy AI, with a spotlight on Federated Learning, data synthesis techniques like dataset distillation and generative methods, large foundation models and related MedTech applications. 
\end{IEEEbiography}

\begin{IEEEbiography}[
{\includegraphics[width=1in,height=1.25in,clip,keepaspectratio]{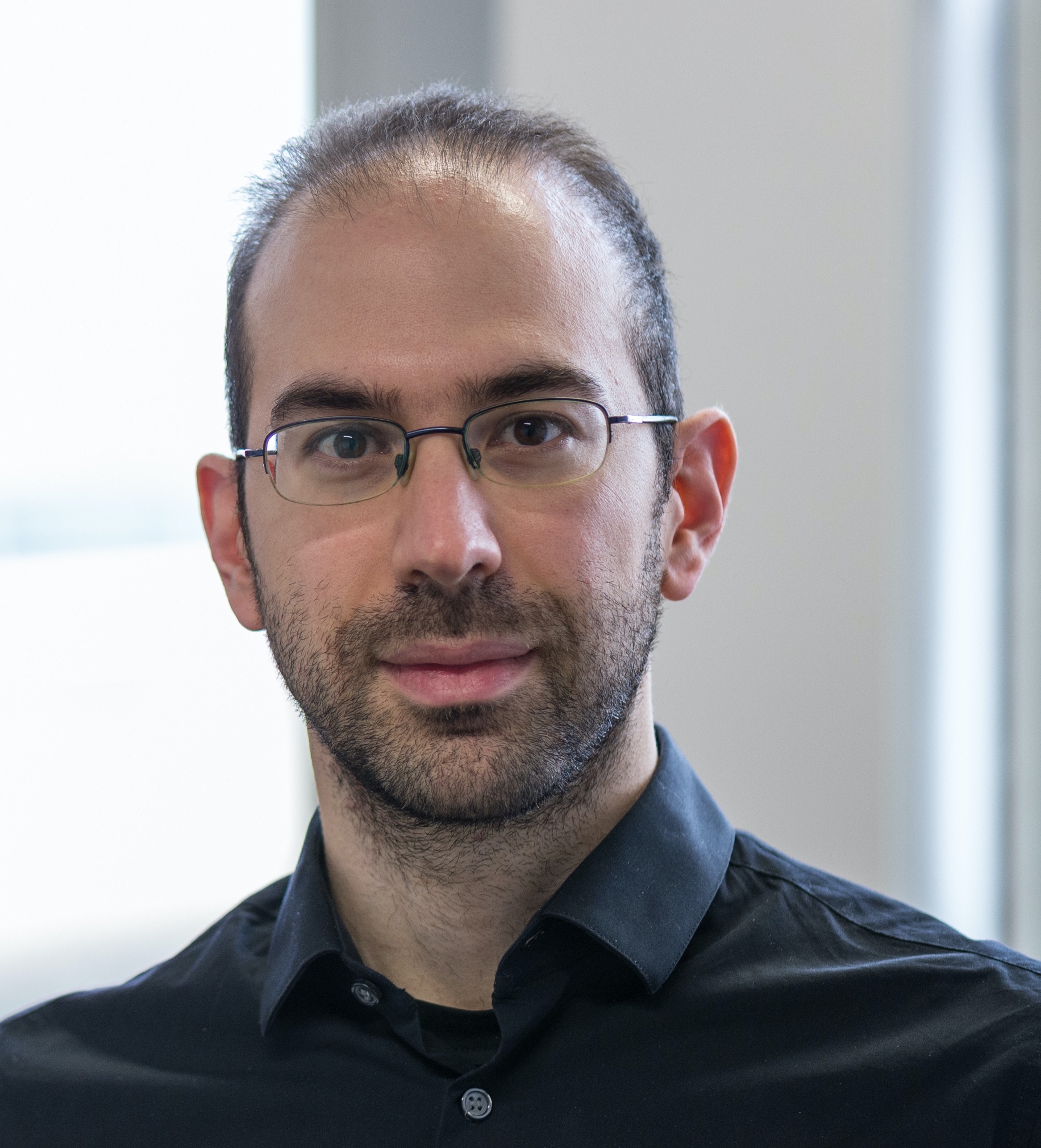}}
]
{Konstantinos G. Kyriakopoulos}~(IEEE Senior Member, IET member)~received the BSc degree in electrical engineering from the Technological Education Institute of Larisa, Greece, in 2003, the MSc degree in digital communication systems, and the PhD degree in computer networks from Loughborough University, Loughborough, UK, in 2004 and 2008, respectively. From 2008 to 2015, he was a Research Associate with the School of Electronic, Electrical, and Systems Engineering, Loughborough University, conducting research in mainly EPSRC projects and successfully licensing research output from his work. Since 2016, he has been an academic member with the Wolfson School of Mechanical, Electronic and Manufacturing Engineering at Loughborough University. His research interests are in the areas of intelligent decision making using machine learning, Evidence Theory, and soft computing techniques. He has extensive experience in computer network security, anomaly detection, contextual awareness, and performance measurements in emerging network paradigms and their applications.
\end{IEEEbiography}

\begin{IEEEbiography}[
{\includegraphics[width=1in,height=1.25in,clip,keepaspectratio]{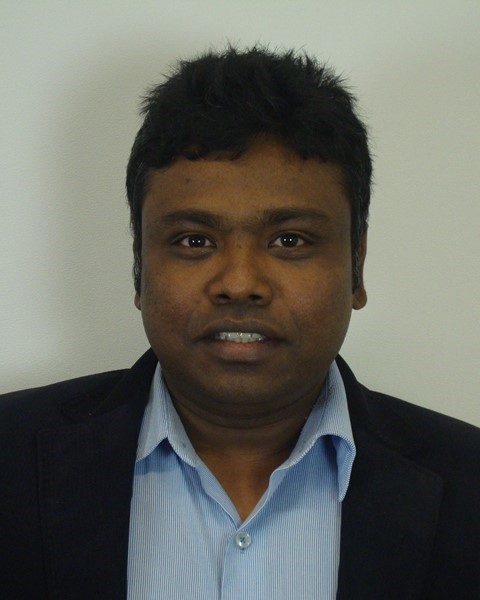}}
]{Sangarapillai Lambotharan} received the Ph.D. degree in signal processing from Imperial College London, U.K., in 1997. He was a Visiting Scientist with the Engineering and Theory Centre, Cornell University, USA, in 1996. Until 1999, he was a Post-Doctoral Research Associate with Imperial College London. From 1999 to 2002, he was with the Motorola Applied Research Group, U.K., where he investigated various projects, including physical link layer modeling and performance characterization of GPRS, EGPRS, and UTRAN. He was with King’s College London and Cardiff University as a Lecturer and a Senior Lecturer, respectively, from 2002 to 2007. He is currently a Professor of Signal Processing and Communications and the Director of the Institute for Digital Technologies, Loughborough University London, U.K. His current research interests include 5G networks, MIMO, blockchain, machine learning, and network security. He has authored more than 250 journal articles and conference papers in these areas. He is a Fellow of IET and Senior Member of IEEE. He serves as a Senior Area Editor for the IEEE Transactions on Signal Processing and as an Associate Editor IEEE Transactions on Communications
\end{IEEEbiography}

\begin{IEEEbiography}[
{\includegraphics[width=1in,height=1.25in,clip,keepaspectratio]{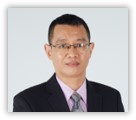}}
]{Qingsong Wei} received the PhD degree in computer science from the University of Electronic Science and Technologies of China, in 2004. He was with Tongji University as an assistant professor from 2004 to 2005. He is a Group Manager and Principal Scientist at the Institute of High Performance Computing, A*STAR, Singapore. His research interests include decentralized computing, federated learning, privacy-preserving AI and Blockchain. He is a senior member of the IEEE.
\end{IEEEbiography}

\end{document}